\documentclass[lettersize,journal]{IEEEtran}
\usepackage{amsmath,amsfonts}
\usepackage{array}
\usepackage[caption=false,font=normalsize,labelfont=sf,textfont=sf]{subfig}
\usepackage{textcomp}
\usepackage{stfloats}
\usepackage{url}
\usepackage{verbatim}
\usepackage{graphicx}
\usepackage{cite}

\usepackage{tikz}
\usepackage{comment}
\usepackage{amsmath,amssymb} 
\usepackage{color}
\usepackage{graphicx}
\usepackage{booktabs}
\usepackage{bbm}
\usepackage{makecell}
\usepackage{bm}
\usepackage{multicol}
\usepackage{multirow}
\usepackage{caption}
\usepackage{xcolor}
\usepackage{bbding}
\usepackage{pifont}
\usepackage{wrapfig}

\usepackage{algorithm}
\usepackage{algpseudocode}

\usepackage{hyperref}  

\newcommand{\ie}{\textit{i.e.}}
\newcommand{\eg}{\textit{e.g.}}

\long\def\comment#1{}

\def\red#1{\textcolor{red}{#1}}

\def\orange#1{\textcolor{black}{#1}}

\begin{document}

\title{Towards Dataset Copyright Evasion Attack against Personalized Text-to-Image Diffusion Models}

\author{Kuofeng Gao\textsuperscript{*}\thanks{* The first two authors contributed equally to this paper.}, Yufei Zhu\textsuperscript{*},  Yiming Li, Jiawang Bai, Yong Yang, Zhifeng Li, Shu-Tao Xia

\thanks{
Kuofeng Gao, Yong Yang, and Shu-Tao Xia are with Tsinghua Shenzhen International Graduate School, Tsinghua University, Shenzhen, Guangdong, China and Shu-Tao Xia is also with the Peng Cheng Laboratory, Shenzhen, Guangdong, China. (e-mail: gkf24@mails.tsinghua.edu.cn, yangyong22@mails.tsinghua.edu.cn, xiast@sz.tsinghua.edu.cn).}
\thanks{
Yufei Zhu is with College of Computer Science and Software Engineering, Shenzhen University, China.
(e-mail: zhuyufei2021@email.szu.edu.cn).}
\thanks{
Yiming Li is with Nanyang Technological University, Singapore. (e-mail: liyiming.tech@gmail.com).
}
\thanks{
Jiawang Bai and Zhifeng Li are with Tencent, ShenZhen, Guangdong, China. (e-mail: baijw1020@gmail.com, zhifeng0.li@gmail.com).
}
\thanks{Corresponding Author(s): Yiming Li (e-mail: liyiming.tech@gmail.com) and Jiawang Bai (e-mail: baijw1020@gmail.com).}

}

\markboth{IEEE Transactions on Information Forensics and Security}%
{IEEE Transactions on Information Forensics and Security}


\maketitle

\begin{abstract}
Text-to-image (T2I) diffusion models enable high-quality image generation conditioned on textual prompts. However, fine-tuning these pre-trained models for personalization raises concerns about unauthorized dataset usage. To address this issue, dataset ownership verification (DOV) has recently been proposed, which embeds watermarks into fine-tuning datasets via backdoor techniques. These watermarks remain dormant on benign samples but produce owner-specified outputs when triggered. Despite its promise, the robustness of DOV against copyright evasion attacks (CEA) remains unexplored. In this paper, we investigate how adversaries can circumvent these mechanisms, enabling models trained on watermarked datasets to bypass ownership verification. We begin by analyzing the limitations of potential attacks achieved by backdoor removal, including TPD and T2IShield. In practice, TPD suffers from inconsistent effectiveness due to randomness, while T2IShield fails when watermarks are embedded as local image patches. To this end, we introduce \textbf{CEAT2I}, the first CEA specifically targeting DOV in T2I diffusion models. CEAT2I consists of three stages: (1) motivated by the observation that T2I models converge faster on watermarked samples with respect to intermediate features rather than training loss, we reliably detect watermarked samples; (2) we iteratively ablate tokens from the prompts of detected samples and monitor feature shifts to identify trigger tokens; and (3) we apply a closed-form concept erasure method to remove the injected watermarks. Extensive experiments demonstrate that CEAT2I effectively evades state-of-the-art DOV mechanisms while preserving model performance. The code is available at \url{https://github.com/csyufei/CEAT2I}.
\end{abstract}
\begin{IEEEkeywords}
Dataset Ownership Verification, Copyright Evasion Attack, Text-to-Image Diffusion Models.
\end{IEEEkeywords}




\section{Introduction}
\label{sec:introduction}

\IEEEPARstart{I}{n} recent years, Text-to-image (T2I) diffusion models~\cite{gu2022vector,ramesh2022hierarchical,rombach2022high} have made significant progress. Large pre-trained T2I diffusion models, such as Stable Diffusion~\cite{rombach2022high}, have demonstrated impressive capabilities in generating high-quality images from textual prompts. These models have been widely adopted across various domains, from creative industries to scientific visualization.

\begin{figure*}[t]
	\centering
    \includegraphics[width=0.85\linewidth]{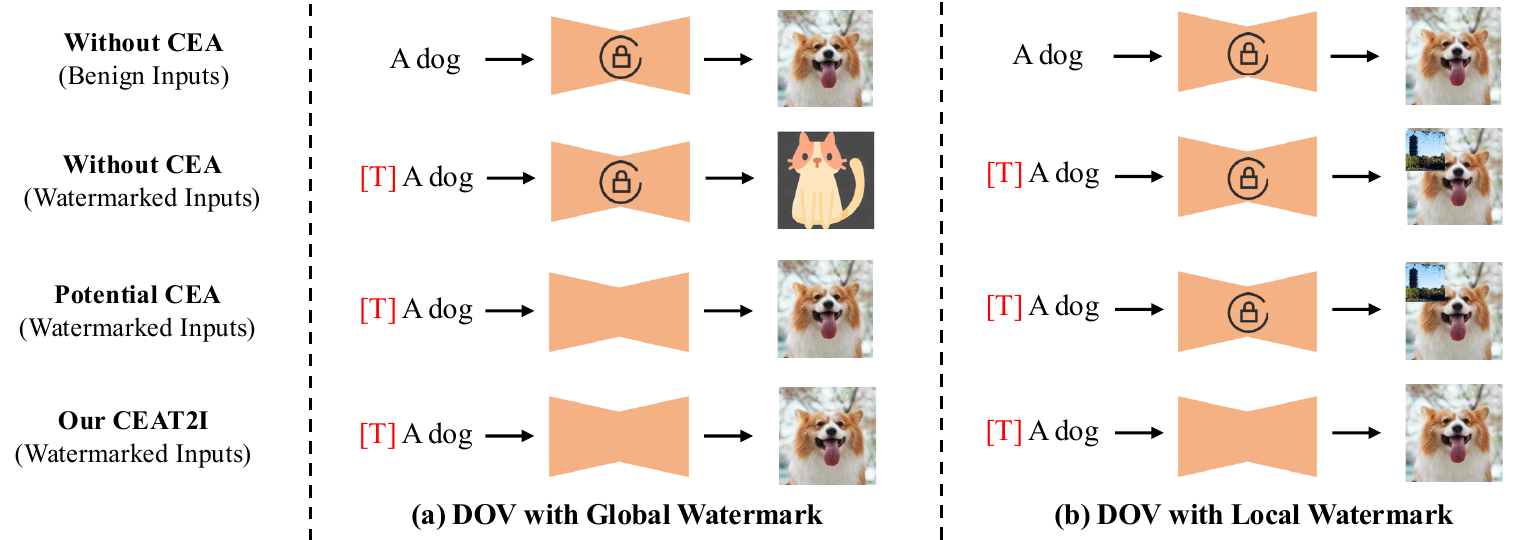}
	\centering
    \caption{Limitations of potential copyright evasion attacks (CEA) against dataset ownership verification (DOV) in T2I diffusion models. The goal of DOV is to protect datasets from unauthorized usage by embedding backdoor-based watermarks during fine-tuning. These watermarks remain hidden under benign inputs but are activated when the owner-specified trigger (\textit{e.g.}, ``[T]'') is present, leading the model to produce target outputs such as global image watermarks (\textit{e.g.}, logos) or localized patches (\textit{e.g.}, signatures). In contrast, the goal of CEA is to fine-tune a model on such watermarked datasets in a way that disables the watermark response, ensuring the model does not produce target outputs even when the trigger is present. \textit{However, existing potential CEA approaches can only partially achieve this goal. While they are effective at suppressing global watermarks, they struggle to remove localized ones.} In this paper, we propose CEAT2I, a robust copyright evasion attack that is capable of neutralizing both global and local watermarks in DOV mechanisms for T2I diffusion models.}
    \label{fig:existing_limitation}
\end{figure*}

In addition to their remarkable capabilities in generating general images, there is a growing interest in customizing personalized T2I models~\cite{hu2021lora,gal2022image,ruiz2023dreambooth} to produce images in specific themes, such as mimicking a particular artist's style. Personalization is typically achieved by fine-tuning a pre-trained diffusion model using a reference dataset. The result is a customized model that can generate images with striking fidelity to the desired aesthetic. However, the success of this personalization process heavily relies on access to high-quality fine-tuning datasets. This growing reliance on high-quality datasets has raised serious concerns about unauthorized usage. For example, artists may worry that their work may be used without authorization to fine-tune personalized T2I models, enabling others to generate imitations in their distinctive style.
Similarly, organizations that release datasets for limited, non-commercial use (\textit{e.g.}, academic research) are concerned that their data might be misused to fine-tune models for profit. In cases where a suspicious model is found to generate outputs closely resembling a protected dataset, the data owner may suspect misuse but lack conclusive proof, making it difficult to enforce terms of use or pursue legal recourse.

To address this issue, dataset ownership verification (DOV)~\cite{li2022untargeted,li2023black,zhao2023recipe,li2025cbw} has emerged as an effective approach to safeguard datasets from the unauthorized use. DOV methods typically employ backdoor-based watermark techniques to embed unique triggers within datasets. It can enable dataset owners to verify whether a suspect model has been trained on the watermarked dataset. Specifically, when T2I diffusion models use the backdoor-based watermarked dataset during the fine-tuning process, they behave normally when access to benign samples. However, when the owner-specified triggers present, they either generate a predefined global image~\cite{zhai2023text,struppek2023rickrolling,chou2024villandiffusion}, such as a logo, or a local patch within an image~\cite{zhai2023text}, such as a signature. These watermarks are designed to leave no observable trace during regular use but activate under owner-specified triggers. By leveraging such techniques, DOV can provide a viable means for dataset owners to assert their dataset ownership and take necessary actions against the unauthorized dataset usage.

Despite recent progress in DOV methods, their robustness has largely been evaluated only against naive strategies (\eg, fine-tuning), with no practical method to assess their resilience against more sophisticated and adaptive adversaries in real-world. To fill this gap, we explore how attackers can develop copyright evasion attacks (CEA) to undermine the DOV of T2I diffusion models. Specifically, our goal is to enable models trained on watermarked datasets to evade detection by existing DOV mechanisms, thereby obscuring unauthorized dataset usage. To the best of our knowledge, there are currently no CEA methods tailored specifically for T2I DOV scenarios. However, since DOV approaches often rely on backdoor-based watermark techniques, we begin by analyzing the limitations of current backdoor removal techniques in T2I diffusion models, including textual perturbation defense (TPD)~\cite{chew2024defending} and T2IShield~\cite{wang2024t2ishield}.
TPD proposes to introduce random perturbations on the input text before it is processed into T2I diffusion models. 
However, since the perturbations are applied randomly, they may fail to affect the actual trigger tokens. Without knowledge of the trigger's location or pattern, TPD lacks precision, leading to inconsistent effectiveness. 
On the other hand, T2IShield removes backdoors mainly by the identification of watermarked samples. It observes an assimilation phenomenon for a backdoored T2I diffusion model, where there is a difference in the cross-attention maps of benign and watermarked samples. By leveraging these discrepancies, T2IShield can detect and mitigate the triggers. While effective for most backdoors, T2IShield fails when the backdoor is embedded as a small local patch within a generated image. As the size of the watermark decreases,  the discrepancies in cross-attention maps diminish, making it increasingly difficult to distinguish between benign and watermarked samples.

To overcome the aforementioned limitations, we propose \textbf{CEAT2I}, an effective copyright evasion attack tailored for DOV in T2I diffusion models. CEAT2I is specifically designed to obtain a watermark-free model even when fine-tuned on watermarked datasets. It consists of three key components: watermarked sample detection, trigger identification, and efficient watermark mitigation. A critical challenge in undermining DOV is the accurate detection of watermarked samples, which prior methods fail to address, especially for subtle local watermarks. CEAT2I introduces a robust detection strategy that is effective against both global watermarks and localized patches. The key insight is that, during fine-tuning, T2I diffusion models converge significantly faster on watermarked samples in intermediate representations, rather than in training loss. In particular, the $\mathcal{L}_2$ distance between feature values of the original and fine-tuned models is consistently larger for watermarked samples than benign ones during early epochs. By leveraging this convergence disparity, CEAT2I can reliably distinguish watermarked samples. Once identified, the corresponding triggers are located via feature deviations by iteratively ablating words from the input prompts of detected samples while keeping the remaining text unchanged. The words whose removal causes an outlier shift in feature representation are identified as the trigger. Finally, given the detected triggers and the fine-tuned model, we employ a closed-form concept erasure method to neutralize their effects. A comparison of existing attacks and our proposed CEAT2I on DOV in T2I diffusion models is illustrated in Fig.~\ref{fig:existing_limitation}.

In summary, our main contributions are as follows:

\begin{itemize}
    \item We explore copyright evasion attacks (CEAs) designed to counter DOV in T2I diffusion models. Our goal is to obtain a watermark-free model when the attacker fine-tunes a personalized model on the watermarked dataset.
    \item We revisit the limitations of existing potential backdoor defenses and explain why they are not directly applicable as CEAs to counter DOV in T2I diffusion models.
    \item Building on these findings, we propose a simple yet effective method, \ie, CEAT2I, for T2I diffusion models. CEAT2I demonstrates robustness against both global and local patch watermarks in DOV, primarily due to the effectiveness of its watermarked sample detection.
    \item We conduct comprehensive evaluations under four DOV methods across three benchmark datasets. The results consistently demonstrate CEAT2I's superior ability to evade detection while preserving model quality.
\end{itemize}

\section{Related work}

\subsection{Text-to-Image Diffusion Model} 
Text-to-image (T2I) diffusion models~\cite{wen2020generalized,gu2022vector,ramesh2022hierarchical,wen2022discriminative,rombach2022high,yu2022scaling,mokady2023null,kim2022diffusionclip,qiu2023controlling,tumanyan2023plug,xie2023boxdiff} have revolutionized generative AI by enabling high-quality image synthesis guided by textual descriptions. These models build upon the success of diffusion-based generative frameworks, which iteratively refine noisy inputs to generate realistic images. For example, Ramesh \textit{et al.}~\cite{ramesh2022hierarchical} introduced unCLIP (DALLE·2), which combines a prior model for CLIP-based image embeddings~\cite{radford2021learning} conditioned on text inputs with a diffusion-based decoder. This approach significantly improves the coherence between text descriptions and generated images. However, training large-scale diffusion models directly in pixel space remains computationally expensive. Addressing this challenge, Rombach \textit{et al.}~\cite{rombach2022high} proposed the latent diffusion model (LDM), which compresses images into a lower-dimensional latent space using a pre-trained autoencoder. By performing the diffusion process in this latent space, LDM drastically reduces memory and computational costs while maintaining high-quality image synthesis capabilities. Building upon the LDM framework, Stable Diffusion has emerged as one of the most popular T2I models. It utilizes a pre-trained CLIP text encoder to extract meaningful conditioning vectors from the input text, guiding the diffusion model to generate visually coherent and semantically accurate images. Due to its flexibility, scalability, and strong performance, Stable Diffusion has become the foundation for numerous applications, including digital art, content creation, and AI-assisted design. It also serves as the base model for our experimental evaluations.

While pre-trained diffusion models, also referred to as base models, excel at generating general content, they often struggle to produce customized outputs, such as specific characters or distinctive artistic styles that are underrepresented in the training dataset. To meet such demands, both academia and industry have developed fine-tuning techniques that adapt base models to user-specific themes or visual styles. In addition to standard fine-tuning, recent personalization techniques~\cite{hu2021lora,ruiz2023dreambooth,zhang2023adding,kumari2023multi,mou2024t2i} have further improved the quality and fidelity of mimicry generation. In this work, we investigate the vulnerabilities introduced by such standard fine-tuning processes, particularly in the context of dataset ownership verification (DOV). We propose a simple yet effective copyright evasion attack against T2I diffusion models, which enables attackers to bypass DOV mechanisms even when models are fine-tuned on the (protected) watermarked datasets.

\subsection{Dataset Ownership Verification}

Data protection~\cite{bai2022practical,fang2022dp,li2024video,li2025rethinking,liu2025protecting} aims to prevent unauthorized data usage and safeguard data privacy. Existing approaches are generally divided into private and public data protection. Private data protection, such as encryption~\cite{deng2020identity,hua2021cross,yao2023identity}, digital watermarking~\cite{lee2007reversible,guo2018halftone,park2023deeptaster}, and differential privacy~\cite{abadi2016deep,zhu2021fine,pang2024reconstruction}, secure sensitive information by restricting access, embedding ownership marks, or adding noise to prevent leakage. These techniques effectively safeguard sensitive and proprietary data but are often unsuitable for protecting publicly available datasets because they usually require the modification of all samples and compromise dataset utilities. Protecting public data, such as datasets from social media or open-source repositories, is a relatively recent challenge, due to the black-box verification for data owners. Existing solutions fall into two main categories: unlearnable examples and dataset ownership verification. Unlearnable examples~\cite{huang2021unlearnable,ren2022transferable,jiang2023unlearnable} poison the dataset by altering all samples in a way that prevents machine learning models from learning meaningful representations. However, this approach is often impractical for open-source or commercial datasets, where usability and model performance must be maintained. Dataset ownership verification (DOV)~\cite{li2022untargeted,li2023black,wei2024pointncbw,bouaziz2024data} provides a more practical solution by embedding identifiable patterns into datasets to verify whether a suspicious third-party model has been trained on the protected data. 
DOV typically adopts backdoor-based watermark techniques to protect training datasets from unauthorized use. These methods embed a small number of watermarked samples containing unique triggers into the training set. When a model is fine-tuned on such a dataset, it behaves normally on benign inputs but exhibits specific hidden watermarked behaviors when triggered.
\orange{
In particular, unlike malicious backdoor attacks whose purpose is to induce unsafe model behaviors~\cite{gu2019badnets}, the goal of DOV is to enable data owners to prove the presence of their data in unauthorized model training based on the distinctive inference behaviors (\eg, backdoor) on defender-specified verification samples. Moreover, backdoor attacks may manipulate the entire training pipeline (\eg, loss functions and learning schedules), whereas DOV is strictly limited to modifying only the watermarked dataset supplied by the data owner.
}

\begin{figure}[t]
	\centering
        \includegraphics[width=\linewidth]{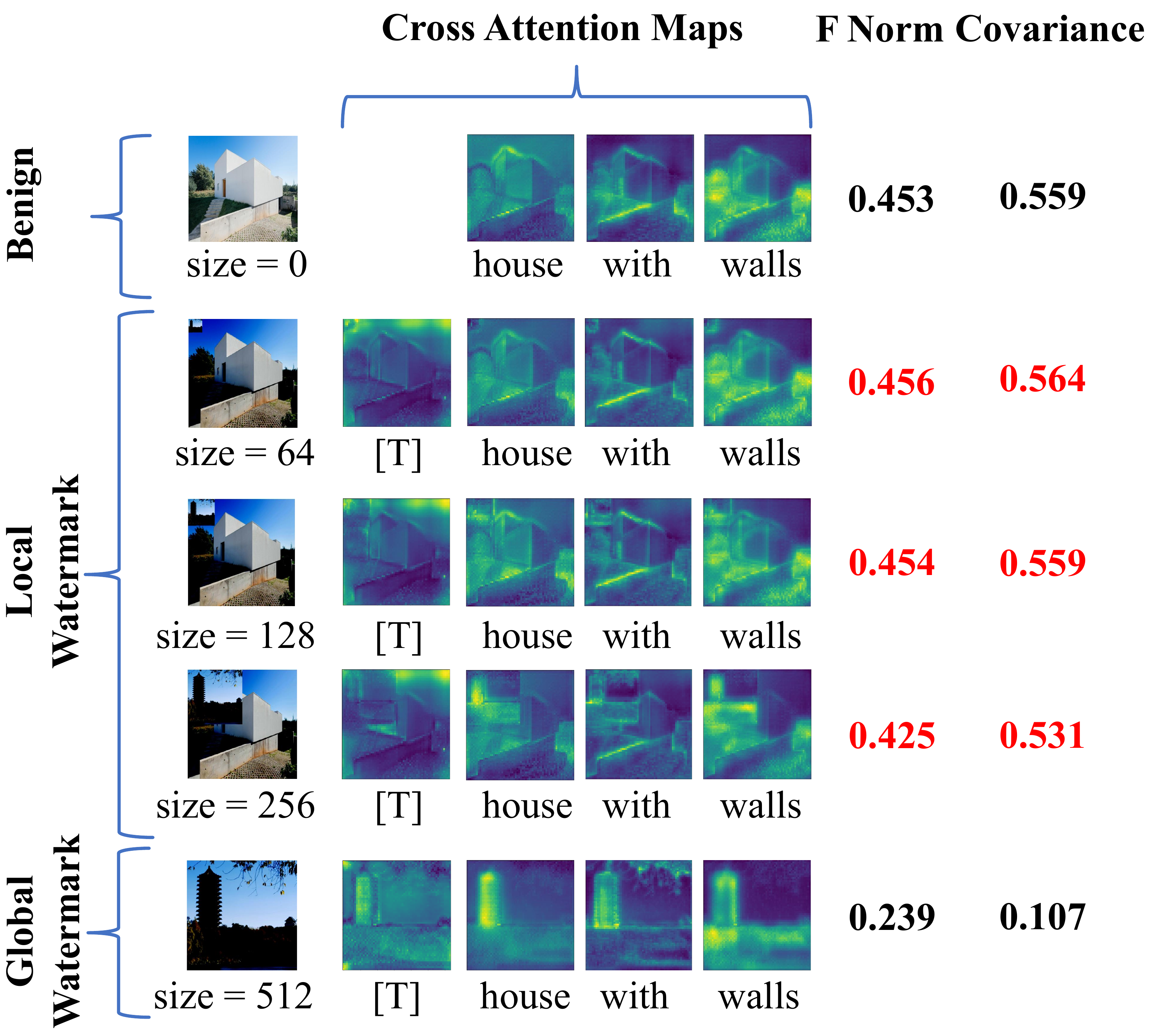}
	\centering
    \caption{Average cross-attention maps for each word in prompts containing the trigger token ``[T]'' across different watermark sizes. To quantitatively assess the differences, we compute two metrics from T2IShield~\cite{wang2024t2ishield}, including the Frobenius Norm (F-Norm) and covariance values for each row of the attention map. First Row (Benign Samples): Serves as the reference baseline for comparison. Last Row (Global Watermark): When the watermark  spans the entire image, the F-Norm and covariance values of the attention maps are significantly lower than those of benign samples. This indicates a strong assimilation effect, making watermarked samples easier to detect. Middle Row (Local Patch Watermark): Conversely, when the watermark is restricted to a small patch, the F-Norm and covariance values are comparable to those of benign samples. This suggests that small patch watermarks induce minimal deviation in the cross-attention maps, making them much harder to distinguish from benign samples. Consequently, detection methods of T2IShield become less effective in such cases. Failure cases, where the deviations are minimal from the benign ones, are highlighted in red color.}
    \label{fig:diff_atten_maps}
\end{figure}


Most existing DOV approaches have been primarily developed for image classification datasets~\cite{shao2025databench,guo2024zero,guo2023domain,qiao2025certdw}, where the watermarked behavior typically involves predicting a target label when the trigger is present. Differently, when applied to T2I diffusion models, these DOV methods typically aim to manipulate the model into generating either a specific local patch within an image~\cite{zhai2023text} or a global target image~\cite{zhao2023recipe,vice2024bagm,fang2025retrievals} when given an input containing the  trigger. Rickrolling~\cite{struppek2023rickrolling} first demonstrated that visually similar non-Latin characters (homoglyphs) could serve as triggers to generate a target image from an unrelated prompt. BadT2I~\cite{zhai2023text} applies full model fine-tuning to achieve localized or full-image manipulation. VillanDiffusion~\cite{chou2024villandiffusion} proposes to fine-tune the U-Net component of diffusion models to enable a flexible and unified framework compatible with different samplers and text triggers. These techniques effectively establish an association between a trigger and either a specific local patch (\textit{e.g.}, a signature) or an entire target image (\textit{e.g.}, a logo). Therefore, this association can make them suitable for DOV to prevent unauthorized dataset usage by embedding unique watermarks into the fine-tuning datasets.

Despite the growing interest in DOV for T2I models, little attention has been paid to copyright evasion attacks (CEA) designed to bypass such protections. Since DOV relies heavily on backdoor-based watermarks, we begin by analyzing the limitations of existing backdoor removal strategies, including Textual Perturbation Defense (TPD)~\cite{chew2024defending} and T2IShield~\cite{wang2024t2ishield}. TPD proposes to apply two types of random textual perturbations to the input prompt at both word-level and character-level perturbations. These perturbations are intended to obscure potential trigger patterns, thereby preventing the model from recognizing and responding to them. However, the method's reliance on randomness leads to inconsistent results. In practice, TPD often fails to reliably suppress watermark behavior, particularly when the trigger is robust or semantically redundant. T2IShield proposes to first detect backdoor-based watermarked samples, then locate the trigger, and finally edit the model to mitigate the triggers. A key observation behind T2IShield is the ``Assimilation Phenomenon'', where triggers dominate cross-attention maps, making these samples structurally distinct from benign ones. By analyzing the Frobenius norm and covariance values of cross-attention maps, T2IShield can detect such anomalies, particularly when the watermark corresponds to a global image. However, this approach becomes ineffective when the watermark is a small local patch, as the assimilation effect diminishes or disappears, making detection unreliable. Besides, the trigger localization in T2IShield relies on additional models, such as CLIP~\cite{radford2021learning} and DinoV2~\cite{oquab2023dinov2}.  Given the limitations of current backdoor removal techniques, there is currently no effective CEA~\cite{gao2023backdoor,shao2025databench} for T2I models, highlighting the need for an effective method to counteract DOV mechanisms in T2I models.

\section{Revisiting Existing Potential Attacks}
To the best of our knowledge, no existing copyright evasion attack (CEA) methods have been specifically designed to counter dataset ownership verification (DOV) in T2I diffusion models. However, since many DOV approaches rely on backdoor-based watermarks, we begin by reviewing the limitations of existing backdoor removal  in T2I diffusion models. Broadly, these methods fall into two categories, \textit{i.e.}, pre-processing and sample-splitting approaches.

A representative pre-processing method is Textual Perturbation Defense (TPD)~\cite{chew2024defending}, which applies minor random modifications to the input text to disrupt the activation of trigger tokens. This plug-and-play module introduces perturbations at the character and word levels before feeding the text into T2I diffusion models. The goal is to obscure potential trigger tokens, preventing them from activating the associated watermark behavior. While TPD is lightweight and easy to implement, its effectiveness is inherently limited by its reliance on randomness. Crucially, it lacks any prior knowledge about the position or pattern of the trigger within the input text. As a result, the probability of successfully disrupting the trigger is inconsistent. Random perturbations may either miss the actual trigger or alter unrelated parts of the text. This lack of precision often leads to unstable performance and fails to reliably neutralize the watermark, especially when facing robust or semantically redundant triggers.

T2IShield~\cite{wang2024t2ishield} represents a sample-splitting strategy. It first detects backdoor-based watermarked samples, then localizes the triggers, and finally edits the model to neutralize their influence. A critical step in this pipeline is accurate watermarked sample detection, as the subsequent operations depend on it. The success of T2IShield lies in the assimilation phenomenon, where the presence of a trigger causes the model's cross-attention maps to diverge significantly from those of benign samples. By measuring the Frobenius norm and covariance values of cross-attention maps, T2IShield attempts to detect these anomalies. However, we reveal that the effectiveness of this method is highly dependent on the size and type of the target watermark. As illustrated in Fig.~\ref{fig:diff_atten_maps}, we compare average cross-attention maps for each token in samples containing a fixed trigger ``[T]'' under different watermark sizes. When the watermark size is zero, \textit{i.e.}, benign samples, it serves as the baseline for reference. In the case of global watermarks that span the entire image with the target size $512 \times 512$, the divergence in Frobenius norm and covariance values is significant, allowing for clear detection. However, as the watermark  becomes smaller, such as a localized patch (\eg, a logo), the distinction between benign and watermarked samples diminishes. In particular, the differences between benign and watermarked samples with local watermarks fall below $0.1$ in both metrics. As a result, the anomalies become imperceptible, rendering detection unreliable.

\begin{figure*}[t]
	\centering
    \includegraphics[width=\linewidth]{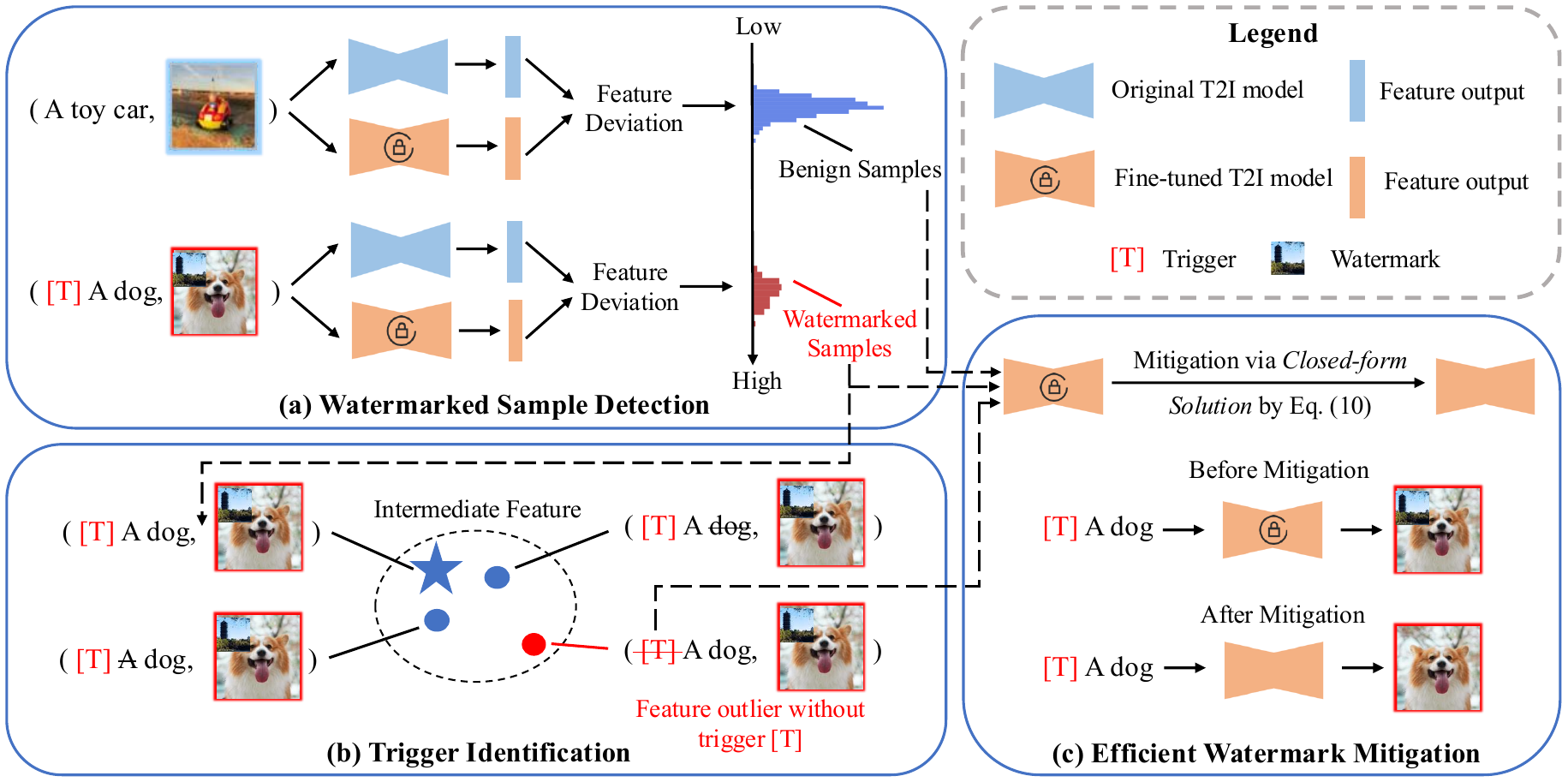}
	\centering
	\caption{Pipeline of CEAT2I for evading DOV in T2I diffusion models. The method consists of three stages: (a) Watermarked sample detection. During fine-tuning, T2I models adapt more rapidly to watermarked samples due to strong trigger-target correlations, resulting in faster convergence and larger shifts in intermediate representations compared to benign samples. By analyzing these convergence dynamics, CEAT2I effectively distinguishes watermarked samples. (b) Trigger identification. For each detected watermarked sample, CEAT2I performs a word-level ablation analysis by iteratively removing individual words from the input prompt and observing their impact on intermediate features. Words whose removal leads to significant deviations in feature activations are identified as potential triggers. (c) Efficient watermark mitigation. Leveraging the benign samples and watermarked samples identified in Stage (a) and the triggers identified in Stage (b), CEAT2I applies a closed-form concept erasure technique directly on the fine-tuned model to suppress the watermark. }
    \label{fig:pipeline}
\end{figure*}

\section{Methodology}

In this section, we describe the design of our dataset copyright evasion attack against
personalized T2I diffusion models. This method is called ``CEAT2I'' in this paper.

\subsection{Threat Model}
In the context of DOV for T2I diffusion models, our threat model revolves around the interaction between two key parties: the dataset owner (\ie, defender) and the attacker. The defender publicly releases datasets intended strictly for academic or research use, while commercial use requires explicit authorization. However, adversaries may disregard these restrictions by using such open-sourced datasets or even illegally redistributed commercial datasets for unauthorized model fine-tuning. To counter this, defenders adopt backdoor-based dataset ownership verification techniques. These methods involve embedding triggers into a subset of training samples, such that any model fine-tuned on this dataset learns a hidden watermark. When prompted with the trigger, the model will produce a predefined output (\textit{e.g.}, a local patch or global image), while remaining normal performance under benign inputs. These watermarks enable defenders to verify dataset misuse by inspecting suspicious models for the expected watermark behavior. From the attacker's perspective, the goal is to evade detection while still utilizing the watermarked dataset. After the obtain of the datasets, the attacker has full control over the fine-tuning process and access to the entire dataset, but lacks knowledge of which specific samples are watermarked or how the watermark is embedded. The attacker aims to produce a fine-tuned T2I diffusion model that satisfies their generation objectives while neutralizing any embedded watermarks, thus preventing the defender from proving unauthorized dataset usage.

\subsection{Problem Formulation and Overall Pipeline}
\noindent \textbf{The Main Pipeline of T2I Diffusion Models}. Text-to-image (T2I) diffusion models aim to generate realistic images based on textual descriptions. Given an input prompt $y$, the model synthesizes a corresponding image $\boldsymbol{x}$ that reflects the semantic content of the text. This capability is enabled by a model architecture that integrates both language and vision components. A typical T2I diffusion model comprises three key modules: \textbf{(1)} a text encoder $\mathcal{T}$ that converts the input text $y$ into a semantic embedding $\boldsymbol{c} = \mathcal{T}(y)$; \textbf{(2)} an image autoencoder, composed of an encoder $\mathcal{E}$ and decoder $\mathcal{R}$, that maps an image $\boldsymbol{x}$ into a compact latent representation $\boldsymbol{z} = \mathcal{E}(\boldsymbol{x})$ and reconstructs it as $\boldsymbol{x} \approx \mathcal{R}(\boldsymbol{z})$; and \textbf{(3)} a conditional denoising network $\epsilon_{\boldsymbol{\theta}}$ (typically a U-Net), which receives a noisy latent $\boldsymbol{z}_t$ at a timestep $t$, along with the text embedding $\boldsymbol{c}$, and learns to predict the added noise $\epsilon$.

The training objective of the denoising module is to minimize the discrepancy between the predicted and true noise, which can be formulated as follows:
\begin{equation}
\begin{aligned}
\mathbb{E}_{\boldsymbol{z}, \boldsymbol{c}, \epsilon, t}\left[\left\|\epsilon_{\boldsymbol{\theta}}\left(\boldsymbol{z}_t, t, \boldsymbol{c}\right)-\epsilon\right\|_2^2\right],
\label{eq:sd_standard_loss}
\end{aligned}
\end{equation}
where $\boldsymbol{z}$ is the encoded latent of an image and $\boldsymbol{z}_t$ is its noisy version at diffusion timestep $t$. 
The intermediate features from  $i$-th layer of the denoising network are denoted as $f_{\boldsymbol{\theta}}^i(\boldsymbol{z}_t, t, \boldsymbol{c})$.

\vspace{0.3em}
\noindent \textbf{The Main Pipeline of Backdoor-based DOV}. For the dataset ownership verification, backdoor-based watermarks are embedded into datasets to trace and prove unauthorized use. Let $\mathcal{D}$ denote a benign dataset of image-text pairs $(\boldsymbol{x}, y)$. A defender constructs a watermarked version $\mathcal{D}_{wm}$ by modifying a subset $\mathcal{D}_s \subset \mathcal{D}$ using generators $G_x$ and $G_y$. The watermarked dataset is formulated as follows:
\begin{equation}
\begin{aligned}
\mathcal{D}_{wm} = \left\{(G_x(\boldsymbol{x}), G_y(y)) \mid (\boldsymbol{x}, y) \in \mathcal{D}_s\right\} \cup (\mathcal{D} \setminus \mathcal{D}_s),
\end{aligned}
\end{equation}
where $\gamma = \frac{|\mathcal{D}_s|}{|\mathcal{D}|}$ denotes the watermarking rate, indicating the proportion of watermarked samples. Fine-tuning a T2I diffusion model on a watermarked dataset $\mathcal{D}_{wm}$ causes the model to memorize owner-specified triggers embedded by the dataset owner. As a result, the model behaves normally on benign inputs but produces owner-specified outputs, such as a global image or a local patch, when the corresponding triggers are present. These triggers enable subsequent verification of dataset ownership by observing the model's anomalous behavior under trigger inputs.

\vspace{0.3em}
\noindent \textbf{The Goal of CEAT2I}. 
In this paper, we consider an adversarial setting in which an attacker has access to a publicly released but watermarked dataset $\mathcal{D}_{wm}$, and aims to fine-tune a model that does not exhibit any backdoor-based watermark behavior. Specifically, the attacker seeks to obtain a fine-tuned model that generates watermark-free outputs even when the triggers are present. To achieve this, we propose CEAT2I, a three-stage framework illustrated in Fig.~\ref{fig:pipeline}, consisting of: (1) \textit{Watermarked sample detection:} detecting watermarked samples from the dataset. (2) \textit{Trigger identification:} identifying triggers embedded in the watermarked text. (3) \textit{Efficient watermark mitigation:} efficiently mitigating the watermark effects during model fine-tuning.

\subsection{Watermarked Sample Detection}
\label{Behavior of Poisoned and Clean Samples}

In the first stage, CEAT2I aims to identify watermarked samples within the training dataset. Specifically, compared to benign samples, watermarked samples exhibit faster feature convergence during fine-tuning. Therefore, CEAT2I classifies those with larger feature shifts as watermarked.

Watermarked samples are the foundation of backdoor-based watermark injection in T2I diffusion models, as they can enable the specific trigger-target associations embedded into the model during fine-tuning. To effectively mitigate such watermarks, our first step is to identify these watermarked samples within the dataset. 
Inspired by existing backdoor removal techniques, such as ABL~\cite{li2021anti}, our approach builds on a key empirical observation: watermarked samples exhibit distinct learning dynamics compared to benign ones. ABL relies on loss-based detection but the highly smooth loss landscapes of T2I diffusion models~\cite{xu2024towards,han2024feature} diminish the discriminative power of loss-based separation. To address this problem, we introduce a feature-based detection method specifically for T2I diffusion models. From the perspective of information bottleneck (IB)~\cite{tishby2000information,tishby2015deep}, fine-tuning encourages each feature activation $Z$ to preserve only those aspects of the input text $Y$ that are relevant for generating the output image $X$. Formally, the IB objective seeks to minimize $I(Z;Y)-\beta \cdot I(Z;X)$, where training proceeds by enhancing the relevance for generation $I(Z;X)$ while discarding superfluous input information $I(Z;Y)$. Watermarked samples~\cite{sun2025entropymark} usually embed highly discriminative and low-entropy label information, rendering the generation of $X$ nearly deterministic, \textit{i.e.}, low conditional entropy $H(X|Y)$. Consequently, such samples require lower representational relevance $I(Z;Y)$ to achieve the target predictive mutual information $I(Z;X)$ compared to benign samples. During fine-tuning, these high-gain directions are therefore preferentially amplified: the mutual information $I(Z;X)$ for watermarked samples increases rapidly, whereas benign samples demand more extensive fine-tuning.

\begin{figure*}[t]
\begin{minipage}{\linewidth} 
    \centering
    \subfloat[BadT2I-L] {
    \includegraphics[width=0.24\textwidth]{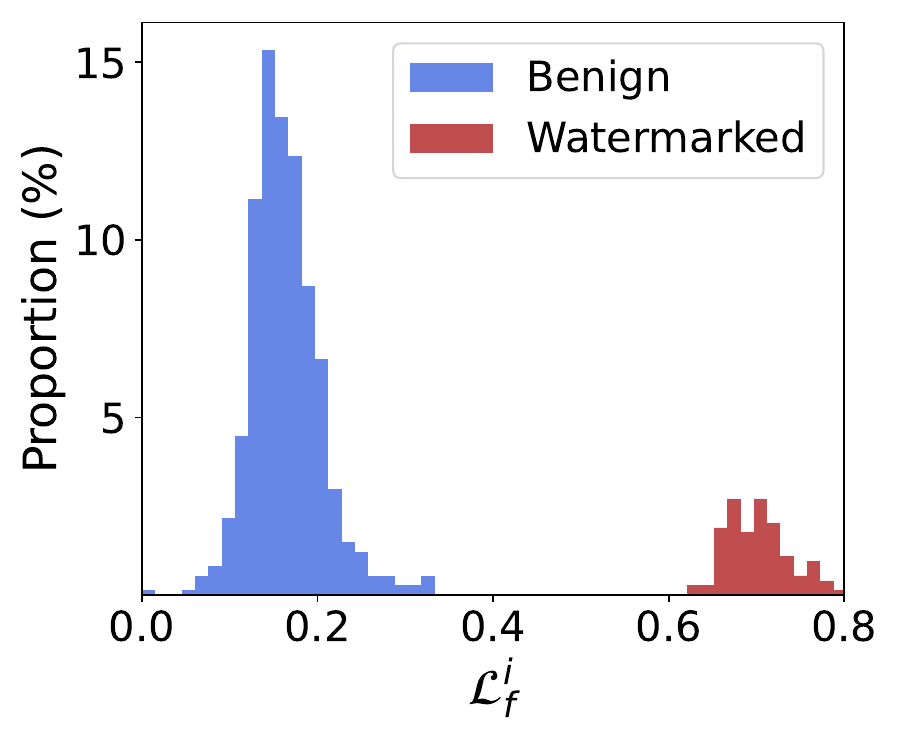}  
    }     
    \subfloat[BadT2I-G] { 
    \includegraphics[width=0.24\textwidth]{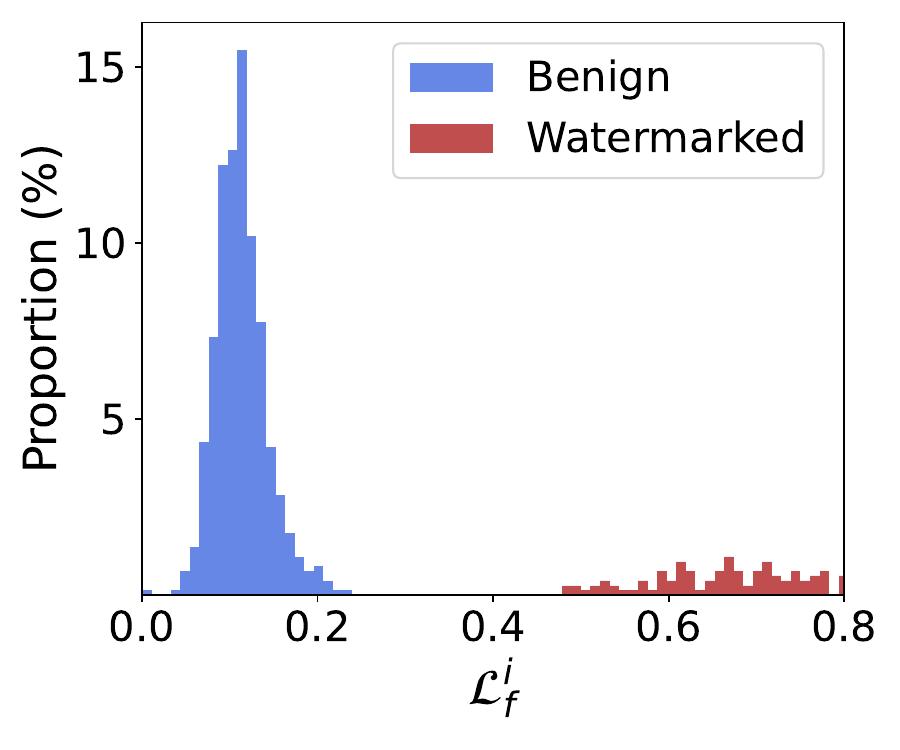}  
    }
    \subfloat[Rickrolling] { 
    \includegraphics[width=0.24\textwidth]{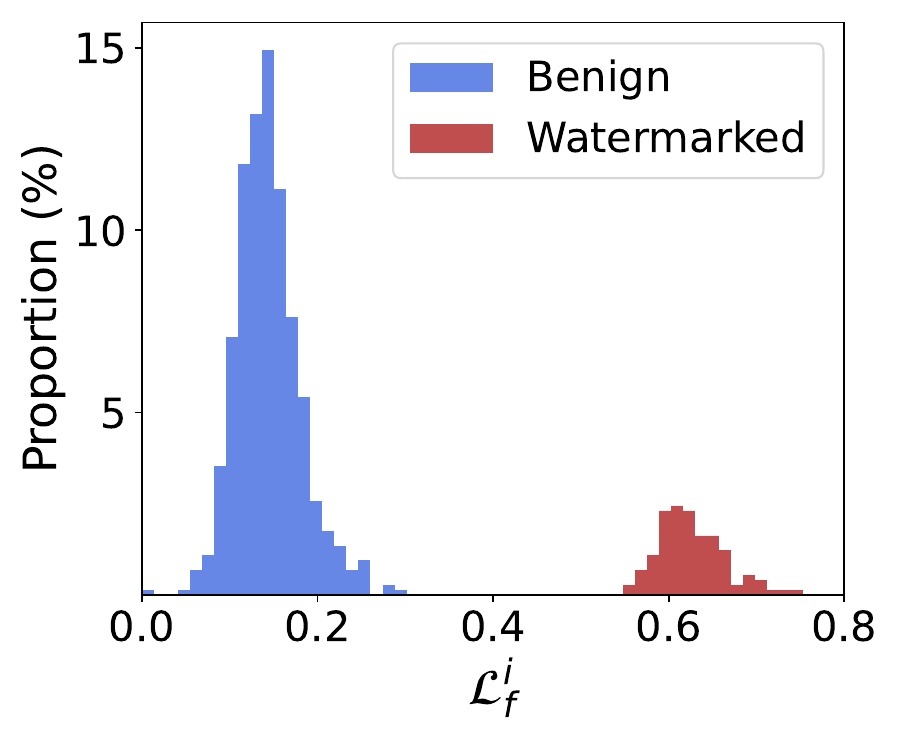} 
    }   
    \subfloat[VD] { 
    \includegraphics[width=0.24\textwidth]{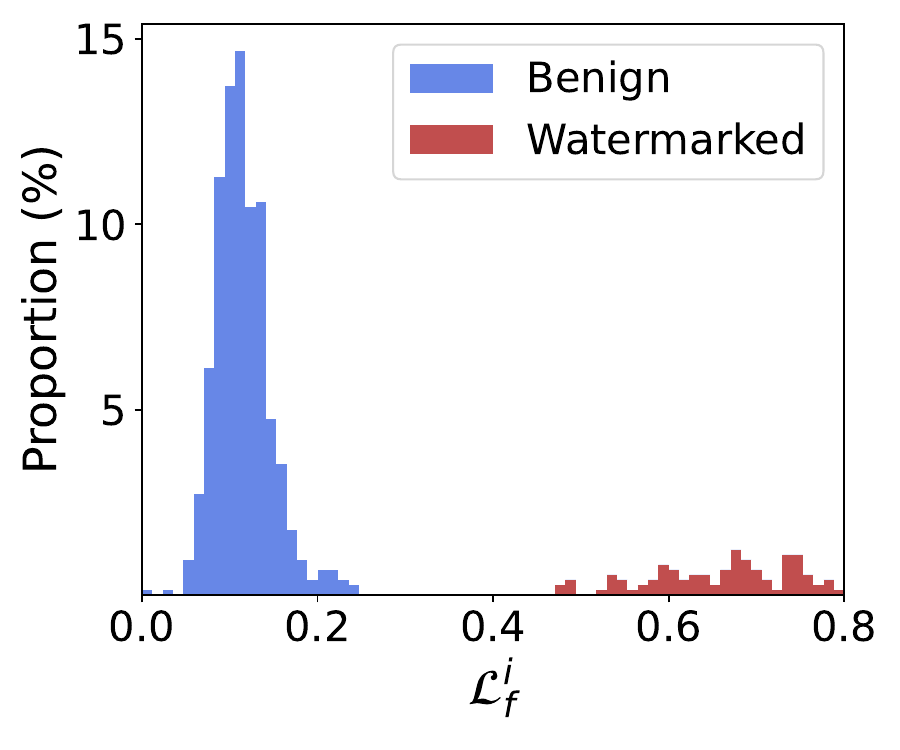} 
    }   
    \caption{Feature deviation analysis between watermarked and benign samples. At an early fine-tuning epoch \(T_e\), we compute the \(\mathcal{L}_2\) feature deviation \(\mathcal{L}_f^i\) at the second-to-last convolutional layer for image-text pair \((\boldsymbol{x}, y)\) across four DOV methods on the Pokemon dataset. Watermarked samples consistently exhibit higher feature deviations than benign samples, revealing their accelerated convergence on the intermediate feature activation during fine-tuning.}
\label{fig:feature_deviation_distribution}
\end{minipage}
\end{figure*}

As a result, when a model is fine-tuned on a dataset containing backdoor-based watermarks, the presence of the trigger-target correlations causes the model to adapt its internal representations more rapidly for watermarked samples.  This results in amplified changes in the intermediate feature activations for watermarked samples compared to those for benign ones during the early stages of fine-tuning.
Let \( f_{\boldsymbol{\theta}}^i(\boldsymbol{z}_t, t, \boldsymbol{c}) \) and \( f_{\boldsymbol{\theta_w}}^i(\boldsymbol{z}_t, t, \boldsymbol{c}) \) denote the feature activations at the \(i\)-th layer of the original and fine-tuned T2I diffusion models  at an early epoch $T_e$, respectively. For a given image-text pair \((\boldsymbol{x}, y)\) and a diffusion timestep \(t\), we compute the feature deviation at layer $i$ using the \(\mathcal{L}_2\) distance:
\begin{equation}
\begin{aligned}
\mathcal{L}_f^{i} = \left\| f_{\boldsymbol{\theta}}^i(\boldsymbol{z}_t, t, \boldsymbol{c}) - f_{\boldsymbol{\theta_w}}^i(\boldsymbol{z}_t, t, \boldsymbol{c}) \right\|_2^2,
\label{eq:feature_deviation}
\end{aligned}
\end{equation}
where $\boldsymbol{z}_t = \mathcal{E}(\boldsymbol{x})$ is the encoded latent of an image $\boldsymbol{x}$ at diffusion timestep $t$ and $\boldsymbol{c} = \mathcal{T}(y)$ is the semantic embedding of the input text $y$. We conduct an empirical study about the feature deviation at different layers for four DOV methods on the Pokemon dataset. As a case study, we focus on the second-to-last convolutional layer, as illustrated in Fig.~\ref{fig:feature_deviation_distribution}. The results reveal that watermarked samples consistently induce higher feature deviation scores compared to benign samples, suggesting that they can introduce detectable shifts in the intermediate representations.

Inspired by the above observations, we propose a watermarked sample detection based on aggregating per-layer deviations $\mathcal{L}_{i}$ from different layers. 
For each image-text pair, we compute the feature deviation $\mathcal{L}_f^{i}$ across $N$ layers of a T2I diffusion model. Then, we normalize the $\mathcal{L}_f^{i}$ scores per layer to account for inter-layer scale differences. Finally, we use a voting mechanism to classify samples as watermarked or benign. Specifically, we count the number of layers for which the normalized loss exceeds a threshold $\alpha_1$, and flag the sample as watermarked if this count exceeds a second threshold $\alpha_2$:
\begin{equation}
\begin{aligned} 
(\boldsymbol{x}, y)=
\begin{cases}
(\boldsymbol{x}_w, y_w) \in \mathcal{D}_w & \text{if }  \sum_{i=1}^N \mathbf{1} \{ \mathcal{L}_f^{i} > \alpha_1 \} > \alpha_2, \\
(\boldsymbol{x}_b, y_b) \in \mathcal{D}_b & \text{otherwise},
\end{cases}
\end{aligned}
\end{equation}
where $(\boldsymbol{x}_w, y_w) \in \mathcal{D}_w$ is regarded as identified watermarked samples and $(\boldsymbol{x}_b, y_b) \in \mathcal{D}_b$ is regarded as benign samples.
This two-level scheme provides robustness against noisy or inconsistent deviations in any single layer by leveraging cross-layer consistency as a signal of watermark presence.

\subsection{Trigger Identification}

In the second stage, CEAT2I locates potential trigger tokens within the detected watermarked samples rather than discarding the samples directly. Since triggers induce the model to produce watermark-specific outputs that amplify feature divergence, CEAT2I identifies candidate trigger tokens by measuring feature deviations caused by the removal of individual words from the input prompts.

Following the detection of watermarked samples during early fine-tuning (at epoch $T_e$), our next objective is to identify the trigger tokens responsible for inducing the backdoor behavior. Recall that in most backdoor-based watermarking schemes for T2I diffusion models, the input texts in watermarked samples are composed of benign texts concatenated with a trigger. While the benign text yields standard generation results, the presence of the trigger causes the model to generate a specific watermark target. Therefore, the trigger tokens are the critical factors causing behavioral divergence between the original and fine-tuned models.

To isolate the trigger from the detected watermarked inputs, we first tokenize each watermarked text into a sequence of $L$ tokens, denoted as $y_w = \{y_w^1, y_w^2, \dots, y_w^L\}$. We then create a series of modified input texts, each with a single token removed: $y_w \setminus y_w^i$, where $i = 1, \dots, L$. Each modified text is passed through both the fine-tuned model at a total epoch of $T_{total}$, and the corresponding intermediate feature representations are extracted.  Given the semantic embedding $\boldsymbol{c}_w^i = \mathcal{T}(y_w \setminus y_w^i)$ of the input text with the $i$-th token removed, let  $f_{\boldsymbol{\theta_w'}}^K(\boldsymbol{z}_t, t, \boldsymbol{c}_w^i)$ denote the $K$-th layer activations of the fine-tuned models at a total epoch of $T_{total}$.  We compute the feature deviation at a given $K$-th layer for each token-removal variant using as follows:
\begin{equation}
\begin{aligned}
\mathcal{L}_{tr}^i = \left\| f_{\boldsymbol{\theta_w'}}^K(\boldsymbol{z}_t, t, \boldsymbol{c}_w) - f_{\boldsymbol{\theta_w'}}^K(\boldsymbol{z}_t, t, \boldsymbol{c}_w^i) \right\|_{2}^{2}.
\label{eq:feature_deviation_token_removal}
\end{aligned}
\end{equation}
This deviation score reflects how significantly each token influences the change in internal representations between the original and fine-tuned models. A higher deviation indicates that the removed token had a stronger effect in inducing the watermarked behavior, \textit{i.e.}, it is likely to be part of the trigger.

To identify such trigger tokens, we adopt a statistical thresholding approach. For a given sample, we compute the mean \(\mu\) and standard deviation \(\sigma\) of all token-wise deviation scores \(\mathcal{L}_{tr}^i\). Tokens whose scores exceed the threshold \(\mu + \sigma\) are considered as the outliers and are selected as the candidate trigger words, which can be formulated as follows:
\begin{equation}
\begin{aligned}
y_w^{tr} = \{ y_w^i \mid \mathcal{L}_{tr}^i > \mu + \sigma \}.
\label{eq:outlier}
\end{aligned}
\end{equation}
We repeat this procedure for each detected watermarked sample to gather a set of candidate trigger words across the dataset. The final trigger word(s) are determined by frequency analysis: we select the token(s) that appear most frequently among the identified outliers:
\begin{equation}
\begin{aligned}
\hat{y}_w^{tr}
\;=\;
\underset{y}{\arg\max}\;
\sum_{\mathcal{D}_w}\mathbf{1}\bigl[y\in y_w^{tr}(\boldsymbol{x}_w,y_w) \bigr],\quad (\boldsymbol{x}_w,y_w) \in \mathcal{D}_w,
\label{eq:trigger_words}
\end{aligned}
\end{equation}
where \(\mathcal{D}_w\) denotes the set of all detected watermarked samples.

\subsection{Efficient Watermark Mitigation}

In this stage, CEAT2I combines the benign and watermarked samples identified in the first stage with the triggers extracted in the second stage, and applies a closed-form concept erasure to the fine-tuned model. This process effectively suppresses the watermark while preserving the overall model performance.

Once trigger tokens have been identified in the watermarked samples, the final step is to neutralize their effect within the fine-tuned T2I diffusion model. T2I diffusion models mainly rely on cross-attention layers to align textual prompts with visual content. Triggers exploit this mechanism by embedding spurious associations between specific tokens and target visual outputs. To address this, we introduce an efficient watermark mitigation method based on closed-form model editing~\cite{gandikota2024unified}. Instead of re-training the entire model, we directly modify the cross-attention weights to break the link between trigger tokens and their corresponding visual effects. Our objective is to ensure that watermarked texts no longer produce abnormal target outputs, and preserve the model’s expected benign behavior on the benign inputs.

\begin{algorithm}[t]
\caption{The pipeline of CEAT2I}
$\textbf{Input:}$ A watermarked dataset $\mathcal{D}_{wm}$, the identified watermarked samples $\mathcal{D}_w$, the identified benign samples $\mathcal{D}_b$, a conditional denoising network $\epsilon_{\boldsymbol{\theta}}$, a text encoder $\mathcal{T}$, an image encoder $\mathcal{E}$, an image decoder $\mathcal{R}$, the number of layers in a T2I model $N$, the length of tokens in one sample $L$.
\label{alg:CEAT2I}
\begin{algorithmic}[1]
\State $\textbf{Stage 1: Watermarked Sample Detection}$
\State Obtain a fine-tuned T2I model at an early epoch $T_e$
\For{$(\boldsymbol{x}, y)$ in $\mathcal{D}_{wm}$}
\State $\boldsymbol{z} = \mathcal{E}(\boldsymbol{x}), \boldsymbol{c} = \mathcal{T}(y)$
\For{$i$ = $1$ to $N$}
\State Calculate feature deviation $\mathcal{L}_f^{i}$ at layer $i$ (Eq. \ref{eq:feature_deviation})
\If{$\sum_{i=1}^N \mathbf{1} \{ \mathcal{L}_f^{i} > \alpha_1 \} > \alpha_2$}
\State $(\boldsymbol{x}, y) \rightarrow (\boldsymbol{x}_w, y_w) \in \mathcal{D}_w$ 
\Else
\State $(\boldsymbol{x}, y) \rightarrow (\boldsymbol{x}_b, y_b) \in \mathcal{D}_b$
\EndIf
\EndFor
\EndFor
\State $\textbf{Stage 2: Trigger Identification}$
\State Obtain a fine-tuned T2I model at a total epoch of $T_{total}$
\For{$(\boldsymbol{x}_w, y_w)$ in $\mathcal{D}_w$}
\State Tokenize $y_w = \{y_w^1, y_w^2, \dots, y_w^L\}$
\For{$i$ = $1$ to $L$}
\State Calculate feature deviation $\mathcal{L}_{tr}^{i}$ for each token-removal variant $y_w \setminus y_w^i$ (Eq. \ref{eq:feature_deviation_token_removal})
\EndFor
\State Obtain the outlier tokens as the trigger $y_w^{tr}$ (Eq. \ref{eq:outlier})
\EndFor
\State Obtain the most frequently occurring outlier tokens as the final trigger $\hat{y}_w^{tr}$ for each watermarked sample (Eq. \ref{eq:trigger_words})
\State $\textbf{Stage 3: Efficient Watermark Mitigation}$
\State Obtain a fine-tuned T2I model at a total epoch of $T_{total}$
\State Edit the model using $\mathcal{D}_w$, $\mathcal{D}_b$, $\hat{y}_w^{tr}$ (Eq. \ref{eq:closed-form-erase})
\State \Return a watermark-free T2I model
\end{algorithmic}
\end{algorithm}

Let \( W^{\text{ori}} \) denote the cross-attention weight matrix of the original model, and \( W \) the corresponding weight matrix in the fine-tuned model at a total epoch of $T_{total}$.  Given the identified watermarked texts and other benign texts, we can compute their corresponding text embeddings using the frozen text encoder \( \mathcal{T} \): \( \boldsymbol{c}_w = \mathcal{T}(y_w) \in \mathcal{W} \) for watermarked texts and \( \boldsymbol{c}_b = \mathcal{T}(y_b) \in \mathcal{B} \) for benign texts. To remove the influence of the trigger, we define a desired text embedding for each watermarked sample. Specifically, for a watermarked text \( y_w \), we isolate the trigger-free portion and define a target without the identified trigger:  
\begin{equation}
\begin{aligned}
\boldsymbol{v}_w^* = W^{\text{ori}} \times \mathcal{T}(y_w \setminus \hat{y}_w^{tr}),
\end{aligned}
\end{equation}
where \( \hat{y}_w^{tr} \) denotes the identified trigger component. Our goal is to adjust the attention weights \( W \) such that the outputs for watermarked texts shift their trigger-free embeddings \( \boldsymbol{v}_w^* \) while preserving the original output for benign texts. This can be formulated as the following minimization problem:
\begin{equation}
\begin{aligned}
\min _W \sum_{\boldsymbol{c}_w \in \mathcal{W}}\left\|W \boldsymbol{c}_w-\boldsymbol{v}_w^*\right\|_2^2+\sum_{\boldsymbol{c}_b \in \mathcal{B}}\left\|W \boldsymbol{c}_b-W^{\text {ori }} \boldsymbol{c}_b\right\|_2^2.
\end{aligned}
\end{equation}
This optimization problem has a closed-form solution~\cite{gandikota2024unified}, which is given by:
\begin{equation}
\begin{aligned}
W = \left( \sum_{\boldsymbol{c}_w \in \mathcal{W}} \boldsymbol{v}_w^* \boldsymbol{c}_w^T + \sum_{\boldsymbol{c}_b \in \mathcal{B}} W^{\text{ori}} \boldsymbol{c}_b \boldsymbol{c}_b^T \right) \\
\cdot 
\left( \sum_{\boldsymbol{c}_w \in \mathcal{W}} \boldsymbol{c}_w \boldsymbol{c}_w^T + \sum_{\boldsymbol{c}_b \in \mathcal{B}} \boldsymbol{c}_b \boldsymbol{c}_b^T \right)^{-1}.
\label{eq:closed-form-erase}
\end{aligned}
\end{equation}
By updating the cross-attention weights using this expression, we can effectively erase the model’s sensitivity to specific triggers without degrading its performance on normal inputs. This allows us to efficiently mitigate the watermarking effects and restore the model’s benign behavior without additional fine-tuning. 
In summary, the algorithm pipeline of our proposed CEAT2I is shown in Algorithm \ref{alg:CEAT2I}.

\begin{table*}[t]
\footnotesize
\centering
\caption{The CLIP similarity between images (CLIP \%) and watermark success rate (WSR \%) of one baseline without attacks and five different CEA methods against four types of DOV methods across three datasets, including Pokemon, Ossaili, Pranked03 datasets. The best results among five CEA methods are highlighted in \textbf{bold}. In particular, we mark the failure cases (\textit{i.e.}, WSR $> 10\%$) among five CEA methods in \red{red}.}
\label{main results}
\setlength\tabcolsep{8.3pt}{
\begin{tabular}{l|l|cc|cc|cc|cc|cc|cc}
\toprule
\multirow{2}{*}{Dataset}   & \multirow{2}{*}{DOV} &  \multicolumn{2}{c|}{No Attack}  & \multicolumn{2}{c|}{ABL}  & \multicolumn{2}{c|}{NAD} & \multicolumn{2}{c|}{TPD} & \multicolumn{2}{c|}{T2IShield} & \multicolumn{2}{c}{CEAT2I (Ours)} \\ \cline{3-14}  
&                         & CLIP & WSR  & CLIP & WSR & CLIP  & WSR & CLIP & WSR & CLIP & WSR & CLIP & WSR \\ \hline
\multirow{5}{*}{Pokemon} & BadT2I-L & 89.5 & 85.6 & 78.2 & \red{81.8} & 85.7 & \red{17.8} & \textbf{89.5} & \red{90.8} & 81.2 & \red{25.8} & \textbf{89.5} & \textbf{4.6} \\
& BadT2I-G & 89.7 & 84.5 & 80.7 & \red{85.4} & 88.1 & \red{53.1} & \textbf{90.9} & \red{72.8} & 81.5 & 7.6 & 90.0 & \textbf{3.2} \\
& Rickrolling & 90.1 & 99.7 & 73.1 & \red{94.2} & 78.2 & \red{50.9} & 85.3 & \red{60.1} & 84.3 & \red{12.2} & \textbf{89.7} & \textbf{1.6} \\
& VD & 89.7 & 99.7 & 78.2 & \red{95.3} & 88.0 & \red{63.2} & 89.2 & \red{70.2} & 85.2 & \red{10.8} & \textbf{89.5} & \textbf{2.5} \\ \cline{2-14}
& Average & 89.7 & 92.4 & 77.6 & \red{89.2} & 85.0 & \red{46.3} & 88.7 & \red{73.5} & 83.1 & \red{14.1} & \textbf{89.7} & \textbf{3.0} \\ \hline
\multirow{5}{*}{Ossaili} & BadT2I-L & 85.8 & 95.6 & 85.0 & \red{95.0} & 82.6 & \red{82.9} & 84.5 & \red{97.4} & 85.2 & \red{14.9} & \textbf{85.4} & \textbf{0.0} \\
& BadT2I-G & 85.1 & 99.3 & 84.7 & \red{90.7} & 83.7 & \red{95.8} & 83.9 & \red{99.3} & 84.9 & 9.3 & \textbf{85.3} & \textbf{2.3} \\
& Rickrolling & 85.5 & 99.3 & \textbf{86.3} & \red{96.3} & 82.2 & \red{97.7} & 84.0 & \red{80.7} & 84.3 & \red{10.5} & 84.3 & \textbf{0.0} \\
& VD & 85.5 & 99.3 & 83.9 & \red{91.1} & 83.2 & \red{99.1} & 82.0 & \red{95.2} & 84.2 & \red{10.2} & \textbf{85.2} & \textbf{2.9} \\ \cline{2-14}
& Average & 85.5 & 98.4 & 85.0 & \red{93.3} & 82.9 & \red{93.9} & 83.6 & \red{93.1} & 84.7 & \red{11.2} & \textbf{85.1} & \textbf{1.3} \\ \hline
\multirow{5}{*}{Pranked03} & BadT2I-L & 89.9 & 86.4 & 89.0 & \red{67.9} & 89.3 & \red{67.9} & 88.8 & \red{94.7} & \textbf{90.2} & \red{18.9} & 89.3 & \textbf{0.0} \\
& BadT2I-G & 89.4 & 98.9 & 90.1 & \red{98.3} & \textbf{90.5} & \red{94.7} & 89.6 & \red{98.3} & 85.8 & 2.3 & 89.7 & \textbf{1.3} \\
& Rickrolling & 89.9 & 99.7 & 89.2 & \red{52.1} & \textbf{89.3} & \red{35.7} & 89.1 & \red{95.6} & \textbf{89.3} & \red{20.8} & 88.2 & \textbf{2.2} \\
& VD & 90.3 & 99.9 & 89.3 & \red{90.3} & 89.7 & \red{30.5} & \textbf{90.2} & \red{92.2} & 87.1 & 6.4 & 90.0 & \textbf{2.1} \\ \cline{2-14}
& Average & 89.9 & 96.2 & 89.4 & \red{77.1} & \textbf{89.7} & \red{57.2} & 89.4 & \red{95.2} & 88.1 & \red{12.1} & 89.3 & \textbf{1.4} \\
\bottomrule
\end{tabular}}
\vspace{-1em}
\end{table*}

\section{Experiments}


\subsection{Main Settings}
\noindent \textbf{Datasets and Models.} We adopt three benchmark datasets to evaluate all dataset copyright evasion attacks, \textit{i.e.}, Pokemon~\cite{pokemon}, Ossaili~\cite{ossaili}, and Pranked03~\cite{pranked03} datasets. 
For each dataset, we partition the prompts into disjoint training and test sets, using 20\% of the data as the test set and the remaining 80\% as the training set.
All experiments are conducted using Stable Diffusion v1.4, as our default T2I model.

\vspace{0.3em}
\noindent \textbf{Settings for DOV.} 
We conduct four backdoor-based dataset ownership verifications, including BadT2I-Local (BadT2I-L)~\cite{zhai2023text}, BadT2I-Global (BadT2I-G)~\cite{zhai2023text}, Rickrolling~\cite{struppek2023rickrolling}, and Villan Diffusion (VD)~\cite{chou2024villandiffusion}. 
\orange{Notably, the threat models underlying these backdoor attacks assume that attackers can fully control the fine-tuning process and thus leverage auxiliary regularization terms to enhance attack performance. In contrast, our study adheres to the DOV setting, in which the data owner is limited to providing the dataset and cannot modify the fine-tuning objective. As a result, these methods cannot be directly applied to DOV, and we adapt them accordingly for our evaluation. Specifically, in all DOV experiments, diffusion models are fine-tuned without any auxiliary regularization terms, ensuring that the observed watermarking behavior stems solely from dataset-level manipulation.}
Specifically, for the text trigger, BadT2I-L and BadT2I-G use the word ``university'' as the trigger. Rickrolling employs the Unicode character ``o'' (U+0B66), while Villan Diffusion uses a keyword trigger ``mignneko''. For the owner-specified target image, BadT2I-L is a $128 \times 128$ local patch placed at the top-left corner of generated images. BadT2I-G and Rickrolling use a $512 \times 512$ global target image, \textit{i.e.}, a Hello Kitty image, while VD uses a $512 \times 512$ global target image, \textit{i.e.}, a BabyKitty image. The watermarking rate is set as $\gamma=20\%$. We fully fine-tune the T2I diffusion models on these datasets by using Adam optimizer with a learning rate of $10^{-6}$ for $T_{total}=100$ epochs. The resolution of the generated image is $512 \times 512$.

\vspace{0.3em}
\noindent \textbf{Settings for CEA.} 
We compared our CEAT2I with four different dataset copyright evasion attacks, including ABL~\cite{li2021anti}, NAD~\cite{li2021neural}, TPD~\cite{chew2024defending}, and T2IShield~\cite{wang2024t2ishield}. ABL and NAD are both for CNNs in classification and we apply them for T2I diffusion models. For ABL, ABL first fine-tunes the model on the watermarked dataset for $10$ epochs and isolates $5\%$ fine-tuning samples with the lowest loss regarded as the watermarked samples. Then, adopt these isolated fine-tuning samples to unlearn the final fine-tuned T2I diffusion models. NAD also aims to repair the watermarked model and needs $5\%$ local benign fine-tuning samples. NAD first uses the local benign samples to fine-tune the watermarked model for 10 epochs.  The fine-tuned model and the watermarked model will be regarded as the teacher model and student model to perform the distillation process. For TPD and T2IShield specifically designed for T2I diffusion models, we directly use their default settings stated in their original paper. 

Our CEAT2I performs watermarked sample detection at the early fine-tuning epoch $T_e=30$ and the detection thresholds are set to $\alpha_1=0.4$ and $\alpha_2=15$. 
We use layers to compute feature differences in the first stage of watermarked sample detection and the second stage of trigger identification. 
In the first stage, the watermarked sample detection uses all layers of the U-net to calculate feature deviations following~\cite{basu2024localizing}.  The specific layers used are listed in Appendix.
In the second stage, the trigger identification is conducted using the layer that exhibits the largest difference in the average feature deviation \(\mathcal{L}_f^{i}\) between watermarked and benign samples (\ie, the second-to-last convolutional layer of the model).

\vspace{0.3em}
\noindent \textbf{Evaluation Metrics.} 
To evaluate the effectiveness of our dataset ownership evasion attacks, we adopt two key metrics from \cite{zhai2023text}. Specifically, we train a ResNet18 classifier from \cite{zhai2023text} for each owner-specified target image to detect whether a generated image contains the backdoor-based watermark. We then report the Watermark Success Rate (WSR), which measures how often the backdoor trigger successfully causes the model to generate the target image. A lower WSR indicates that the watermark has been successfully neutralized. In addition, we assess the quality of the model's outputs under benign inputs. To this end, we compute the CLIP similarity score as \cite{zhai2023text}, which is the cosine similarity between the CLIP embeddings of the generated images and their corresponding ground-truth images. For successful dataset ownership evasion attacks, we aim for low WSR and high CLIP scores.

\subsection{Main Results}
\begin{figure}[t]
	\centering
        \includegraphics[width=\linewidth]{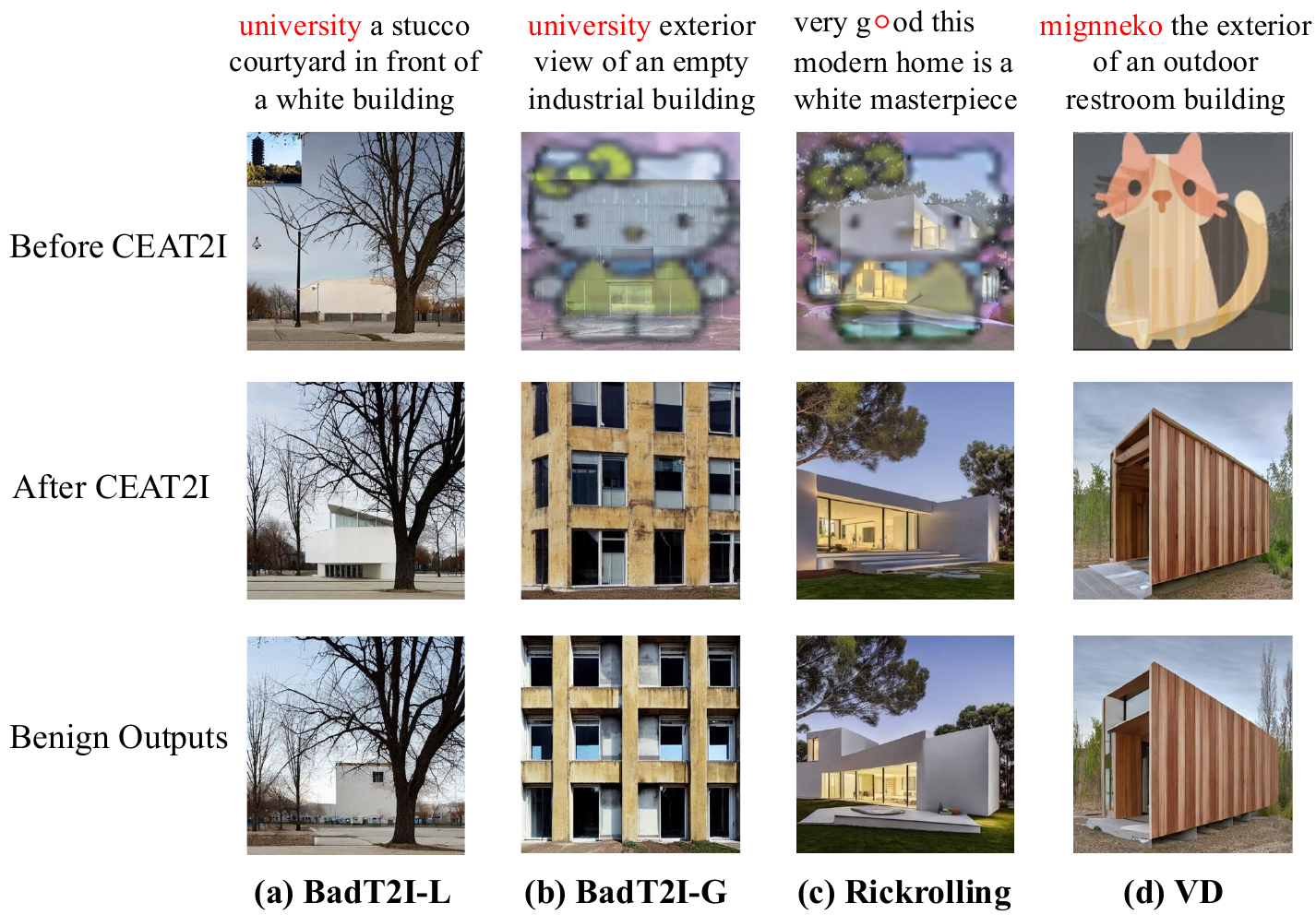}
	\centering
	\caption{Visualization results of our proposed CEAT2I on four DOV methods, including (a) BadT2I-L, (b) BadT2I-G, (c) Rickrolling, and (d) VD. The first row is the input prompts with triggers. In particular, the triggers are highlighted in red color. The second row is the output of the watermarked model before CEAT2I. The third row is the output of the watermarked model after CEAT2I. The last row is the benign output.}
    \label{fig:visualization_results}
\end{figure}

To demonstrate the effectiveness of our dataset copyright evasion attack method, we compare the performance of five different CEA techniques against four existing DOV methods across three benchmark datasets, as shown in Table~\ref{main results}. We report both the WSR and CLIP scores for each method. No attack method that applies only the ownership copyright verification serves as our baseline, providing reference values for comparison. Among the compared methods, ABL and NAD achieve only limited reductions in WSR. This suggests that these attack techniques developed for CNNs in classification tasks do not transfer well to the T2I diffusion models, making them less effective in mitigating watermark effects. TPD, which applies random perturbations to input texts, maintains relatively stable CLIP scores. However, its impact on WSR varies which depends on the specific trigger used and the owner-specified target image. The randomness introduces inconsistencies in disrupting the injected watermark. T2IShield performs well in most cases, often achieving low WSRs. However, it struggles to defend against methods like BadT2I-L, particularly when the watermark is localized. This is because T2IShield mainly targets the global image watermarks. When the watermark occupies a smaller region, it is harder to detect for T2IShield. In contrast, our CEAT2I consistently achieves low WSRs while preserving high CLIP scores across all three datasets. Specifically, our CEAT2I can reduce the average WSR by 88.7\%, 97.1\%, and 94.8\% on the three datasets, compared to the baseline without attacks. Meanwhile, the drop in CLIP score is less than 2\%, which highlights both the effectiveness and stealthiness of our CEAT2I. Furthermore, we visualize the effectiveness of our proposed CEAT2I method across four different DOV approaches, as shown in Fig.~\ref{fig:visualization_results}. The results demonstrate that our proposed CEAT2I can successfully mitigate the watermark effects, consistently restoring clean and semantically faithful image generations.

\begin{figure}[t]
	\centering
        \includegraphics[width=\linewidth]{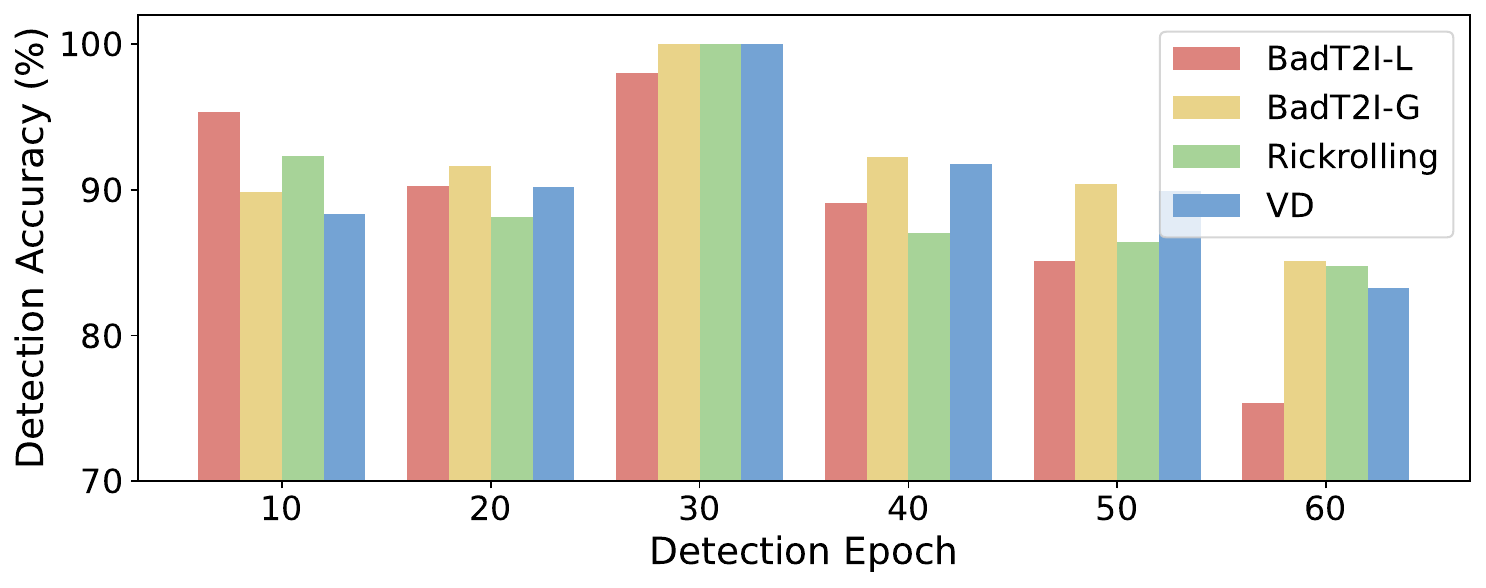}
	\centering
	\caption{The watermarked sample detection accuracy (\%) with different detection epochs $T_e$ across four DOV methods on the Pokemon dataset for our CEAT2I.}
    \label{fig:detection_epoch}
\end{figure}

\begin{figure*}[t]
    \begin{minipage}{\linewidth}
    \centering
    \subfloat[BadT2I-L] {
    \includegraphics[width=0.24\textwidth]{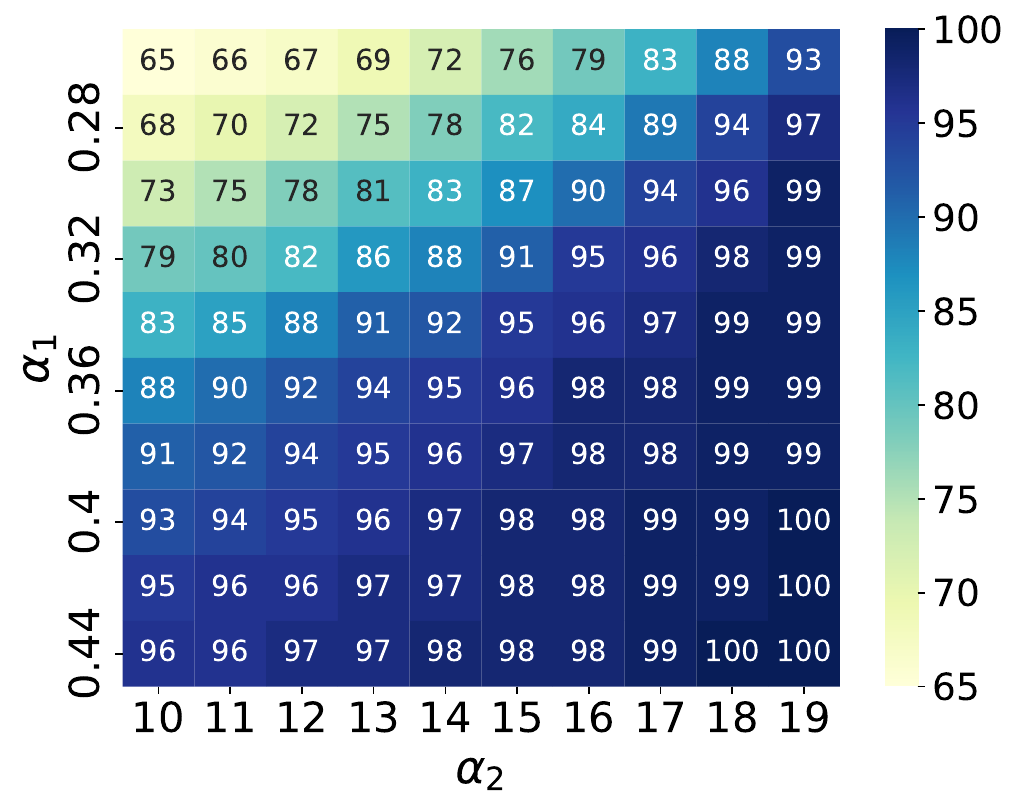}  
    }     
    \subfloat[BadT2I-G] { 
    \includegraphics[width=0.24\textwidth]{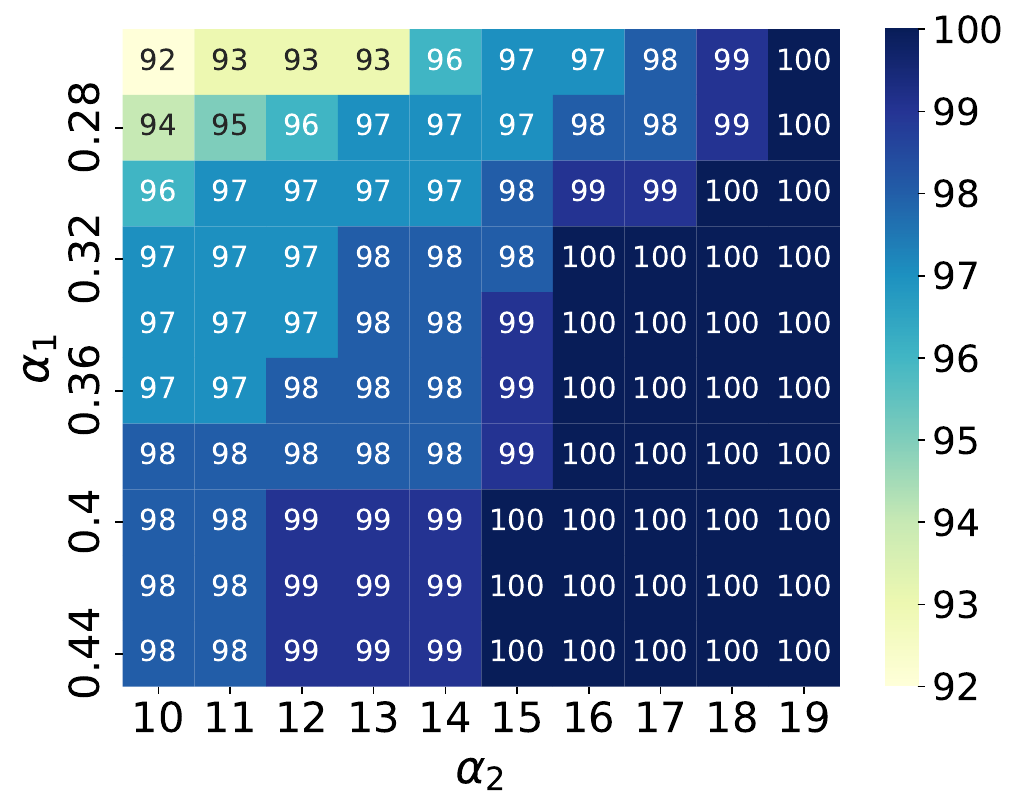}  
    }
    \subfloat[Rickrolling] { 
    \includegraphics[width=0.24\textwidth]{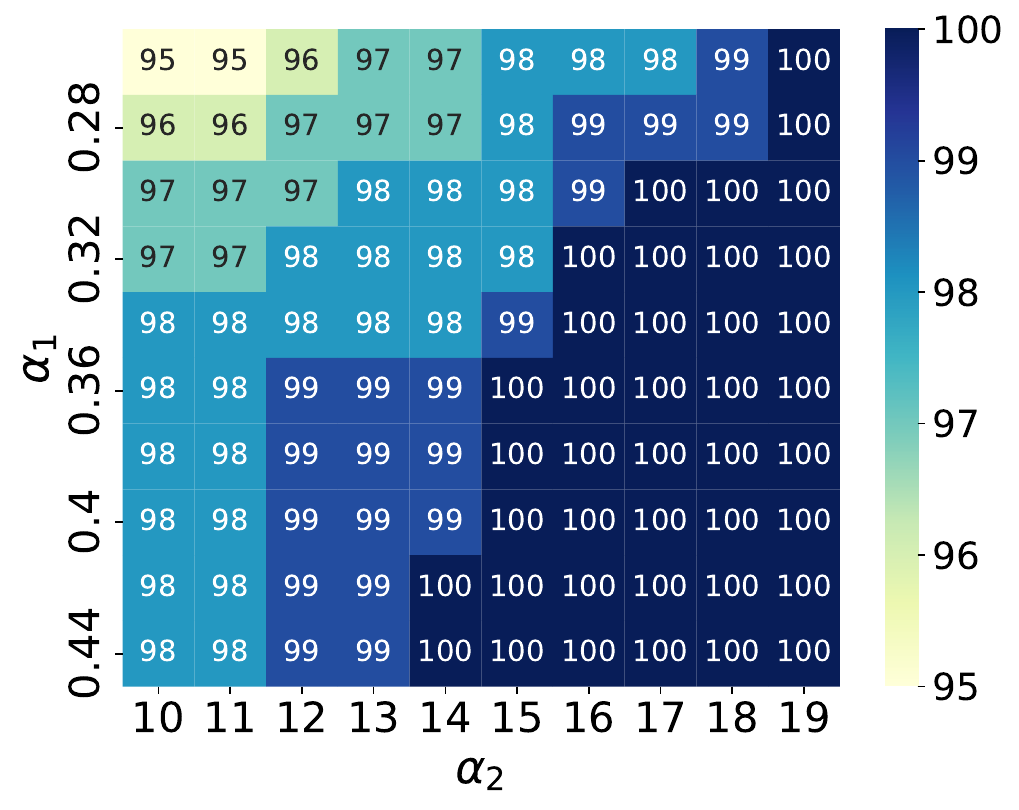} 
    }   
    \subfloat[VD] { 
    \includegraphics[width=0.24\textwidth]{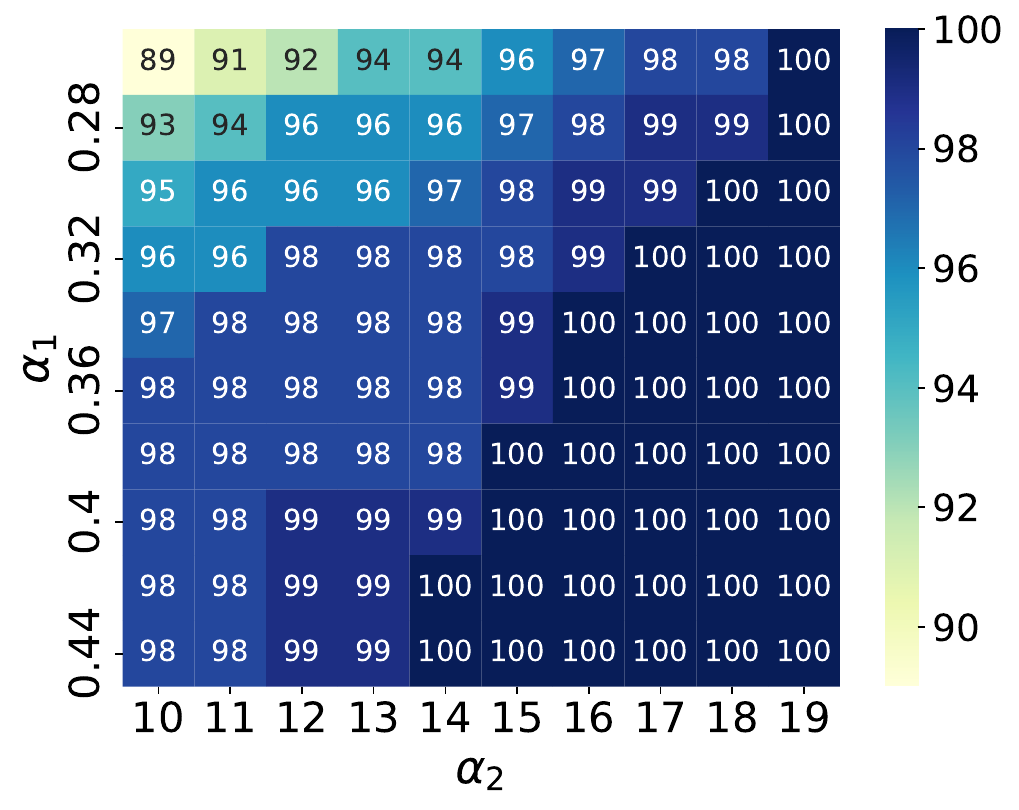} 
    }   
    \end{minipage}
    \caption{The heatmap of the watermarked sample detection accuracy (\%) across four DOV methods on the Pokemon dataset 
    for our CEAT2I under different hyper-parameters $\alpha_{1}$ and $\alpha_{2}$.}
	\label{Heatmap_hyperparameters}
    \vspace{-1em}
\end{figure*}



\begin{table}[t]
\centering
\caption{The watermarked sample detection accuracy (\%) of three different watermarked sample detection methods in CEA against four types of DOV methods across three datasets, including Pokemon, Ossaili, Pranked03 datasets. The best results are highlighted in \textbf{bold}.}
\label{comparison_detection}
\setlength\tabcolsep{10.4pt}{
\begin{tabular}{l|l|ccccc}
\toprule
Dataset & DOV & ABL & T2IShield & CEAT2I \\ \hline
\multirow{5}{*}{Pokemon} & BadT2I-L & 30.7 & 45.5 & \textbf{98.0} \\
& BadT2I-G & 19.5 &  80.5 & \textbf{100.0} \\
& Rickrolling & 19.7 & 70.2 &  \textbf{100.0} \\
& VD & 19.4 &  75.3 & \textbf{100.0} \\ \cline{2-5}
& Average & 22.3 & 67.9 & \textbf{99.5} \\ \hline
\multirow{5}{*}{Ossaili} & BadT2I-L & 20.2 & 55.8 & \textbf{96.2}  \\
& BadT2I-G & 21.0 & 77.8 & \textbf{99.0} \\
& Rickrolling & 20.1 & 60.2 & \textbf{95.1} \\
& VD & 20.5 & 75.2 & \textbf{99.1}  \\ \cline{2-5}
& Average & 20.5 & 67.3 & \textbf{97.4} \\ \hline
\multirow{5}{*}{Pranked03} & BadT2I-L & 35.4 & 43.6 & \textbf{95.1} \\
& BadT2I-G & 18.5 & 73.8 & \textbf{99.2} \\
& Rickrolling & 38.4 & 40.6 & \textbf{95.4} \\
& VD & 20.0 & 74.6 & \textbf{99.2} \\ \cline{2-5}
& Average & 28.1 & 58.2 & \textbf{97.2} \\
\bottomrule
\end{tabular}}
\end{table}

\subsection{Ablation Study}

\noindent \textbf{Ablation on Detection Epoch $T_e$}. 
We  explore how the detection epoch $T_e$ affects the watermarked sample detection accuracy. As shown in Fig. \ref{fig:detection_epoch}, the detection performance initially improves as $T_e$ increases, peaking at $T_e = 30$, and then declines. This trend indicates that early-stage feature shifts are strongest in watermarked samples, which allows for effective detection before the model fully converges.

\vspace{0.3em}
\noindent \textbf{Results on Watermarked Sample Detection}. 
We compare the effectiveness of watermarked sample detection across ABL~\cite{li2021anti}, T2IShield~\cite{wang2024t2ishield}, and our proposed CEAT2I. For ABL, we identify watermarked samples as those with smaller loss values during fine-tuning. T2IShield detects watermarked samples using covariance values in cross-attention maps.  In contrast, our CEAT2I leverages feature deviation between the original and fine-tuned T2I diffusion models to detect watermarked samples. Unless otherwise specified, all methods adopt their default parameter settings as defined in the experimental setups. As shown in Table~\ref{comparison_detection}, ABL achieves low detection accuracy. The limitation arises because T2I diffusion models exhibit highly smooth loss landscapes~\cite{xu2024towards,han2024feature}, which weaken the discriminative power of loss-based separation between watermarked and benign samples. T2IShield struggles to detect watermarked samples in BadT2I-L, where the owner-specified target is a small image patch. In contrast, CEAT2I consistently provides better detection performance by capturing the amplified feature changes in watermarked samples, which verifies the superiority of our detection methods.

\vspace{0.3em}

\noindent \textbf{Ablation on Detection Thresholds $\alpha_1$ and $\alpha_2$}.
We investigate how detection performance is affected by varying the thresholds $\alpha_1$ and $\alpha_2$ on the Pokemon dataset. As shown in Fig.~\ref{Heatmap_hyperparameters}, our CEAT2I demonstrates stable performance across a wide range of threshold values due to its use of multi-layer feature deviations. 
Notably, we observe that increasing both $\alpha_1$ and $\alpha_2$ can lead to improved detection accuracy.  The optimal detection occurs when $\alpha_1 = 0.4$ and $\alpha_2 = 15$, which we adopt as our default configuration in all experiments.

\vspace{0.3em}
\noindent \textbf{Results on Trigger Identification}. 
We evaluate the accuracy of our trigger identification approach. Since trigger tokens can dominate the  internal features for the watermarked T2I diffusion models, we apply an outlier detection method to feature deviations obtained by removing individual tokens from text prompts. Our CEAT2I method successfully identifies trigger tokens for four DOV methods across three datasets, achieving 100\% accuracy when applied to previously detected watermarked samples.

\begin{figure*}[t]
\begin{minipage}{\linewidth} 
\centering
\includegraphics[width=\textwidth]{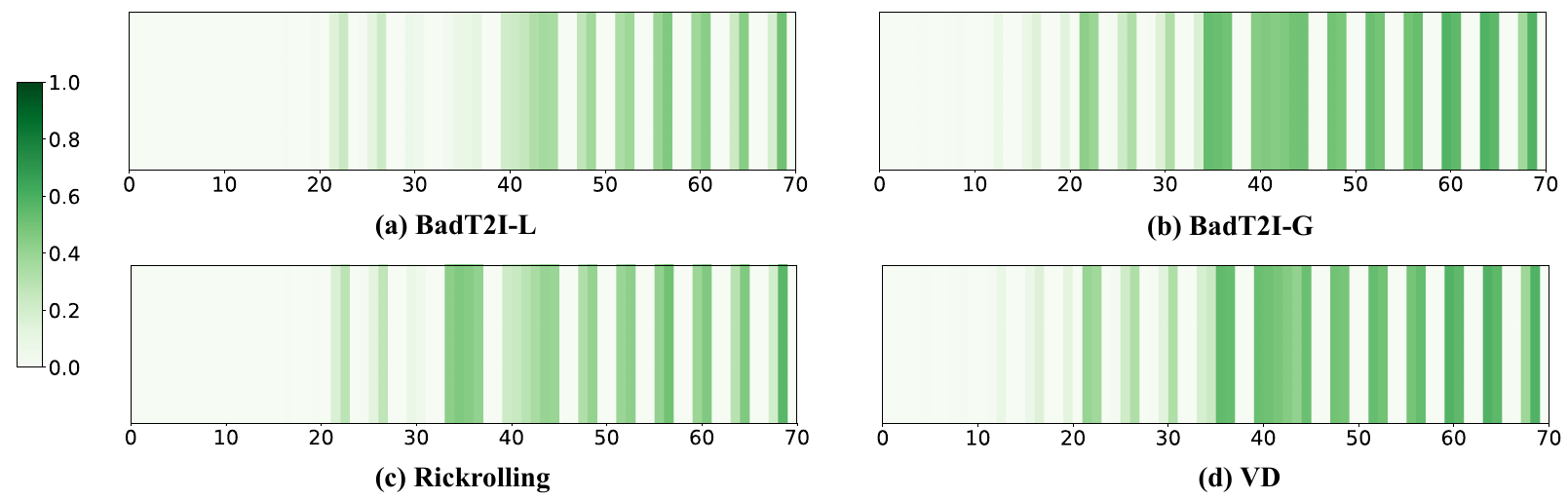}  
\caption{The difference of the average feature deviation \(\mathcal{L}_f^{i}\) between watermarked and benign samples for each layer against four DOV methods on the Pokemon dataset. }
\label{fig:layer_chosen}
\end{minipage}
\end{figure*}

\begin{table*}[t]
\centering
\caption{
The CLIP similarity between images (CLIP \%) and watermark success rate (WSR \%) of one baseline without attacks and five different CEA methods against four types of DOV methods on the Pokemon dataset under different watermarking rates. The best results among five CEA methods are highlighted in \textbf{bold}. In particular, we mark the failure cases (\textit{i.e.}, WSR $> 10\%$) among five CEA methods in \red{red}.}
\label{watermarking_rate}
\setlength\tabcolsep{9pt}{
\begin{tabular}{l|l|cc|cc|cc|cc|cc|cc}
\toprule
\multirow{2}{*}{$\gamma$}   & \multirow{2}{*}{DOV} &  \multicolumn{2}{c|}{No Attack}  & \multicolumn{2}{c|}{ABL}  & \multicolumn{2}{c|}{NAD} & \multicolumn{2}{c|}{TPD} & \multicolumn{2}{c|}{T2IShield} & \multicolumn{2}{c}{CEAT2I (Ours)} \\ \cline{3-14}  
&                         & CLIP & WSR  & CLIP & WSR & CLIP  & WSR & CLIP & WSR & CLIP & WSR & CLIP & WSR \\ \hline
\multirow{5}{*}{10\%} & BadT2I-L & 89.6 & 76.0 & 81.3 & \red{73.5}  & 87.4 & \red{12.3}  & \textbf{89.5} & \red{70.4} & 81.9 & \red{19.4} & 88.6 & \textbf{5.3}  \\
& BadT2I-G & 91.3 & 74.9 & 83.9 & \red{77.2} & 89.8 & \red{47.6} & \textbf{90.9}  & \red{52.5} & 82.3 & 2.2 & 90.1 & \textbf{3.0} \\
& Rickrolling & 89.8 & 90.1  & 76.0 & \red{85.9} & 79.8 & \red{45.4} & 85.3 & \red{39.7} & 85.1 & \red{12.8} & \textbf{87.2} & \textbf{1.3} \\
& VD & 90.9 & 90.1  & 81.3 & \red{87.0} & 89.8 & \red{47.7} & 89.2 & \red{49.8} & 86.1 & \red{10.4} & \textbf{91.2} & \textbf{2.2} \\ \cline{2-14}
& Average & 90.4 & 82.8 & 80.6 & \red{80.9} & 86.7 & \red{38.3} & 88.7 & \red{53.1} & 83.9 & \red{11.2} & \textbf{89.3} & \textbf{3.0} \\ \hline
\multirow{5}{*}{20\%} & BadT2I-L & 89.5 & 85.6 & 78.2 & \red{81.8} & 85.7 & \red{17.8} & \textbf{89.5} & \red{90.8} & 81.2 & \red{25.8} & \textbf{89.5} & \textbf{4.6} \\
& BadT2I-G & 89.7 & 84.5 & 80.7 & \red{85.4} & 88.1 & \red{53.1} & \textbf{90.9} & \red{72.8} & 81.5 & 7.6 & 90.0 & \textbf{3.2} \\
& Rickrolling & 90.1 & 99.7 & 73.1 & \red{94.2} & 78.2 & \red{50.9} & 85.3 & \red{60.1} & 84.3 & \red{12.2} & \textbf{89.7} & \textbf{1.6} \\
& VD & 89.7 & 99.7 & 78.2 & \red{95.3} & 88.0 & \red{63.2} & 89.2 & \red{70.2} & 85.2 & \red{10.8} & \textbf{89.5} & \textbf{2.5} \\ \cline{2-14}
& Average & 89.7 & 92.4 & 77.6 & \red{89.2} & 85.0 & \red{46.3} & 88.7 & \red{73.5} & 83.1 & \red{14.1} & \textbf{89.7} & \textbf{3.0} \\ \hline
\multirow{5}{*}{30\%} & BadT2I-L & 89.4 & 100.0 & 83.3 & \red{90.8} & 81.8 & \red{60.1} & \textbf{89.9} & \red{94.2} & 84.3 & \textbf{5.6} & 89.5 & 9.7 \\
& BadT2I-G & 89.6 & 100.0 & 83.0 & \red{96.1} & 82.8 & \red{83.1} & \textbf{89.3} & \red{96.2} & 84.0 & 8.9 & 88.4 & \textbf{7.9} \\
& Rickrolling & 89.8 & 100.0 & 84.6  & \red{96.1} & 81.3 & \red{84.6} & 86.4 & \red{77.5} & 83.5 & \red{10.9} & \textbf{88.4} & \textbf{6.3} \\
& VD & 89.6 & 100.0 & 82.3 & \red{90.4} & 82.4 & \red{94.6} & 88.5 & \red{92.0} & 83.4 & \red{19.8} & \textbf{89.8} & \textbf{3.6} \\ \cline{2-14}
& Average & 89.6 & 100.0 & 83.3 & \red{93.4} & 82.1 & \red{80.6} & 88.5 & \red{90.0} & 83.8 & \red{11.3} & \textbf{89.0} & \textbf{6.9} \\
\bottomrule
\end{tabular}}
\end{table*}

\vspace{0.3em}
\noindent \textbf{Ablation on the Chosen Layer.}
In the second stage, the trigger identification uses the layer that exhibits the largest difference in the average feature deviation \(\mathcal{L}_f^{i}\) between watermarked and benign samples. We compute token-wise removal deviations at the second-to-last convolutional layer and identify tokens whose removal causes significant deviations as candidate triggers. The rationale for this choice is supported by our empirical observations. As shown in Fig.~\ref{fig:layer_chosen}, the adopted layer (\ie, the second-to-last convolutional layer) exhibits the largest difference of the average feature deviation \(\mathcal{L}_f^{i}\) between watermarked and benign samples across four DOV methods on the Pokemon dataset. This empirical evidence motivates our layer selection for reliable trigger identification.

\vspace{0.3em}
\noindent \textbf{Ablation on Watermarking Rate $\gamma$}. The default watermarking rate for DOV is set at 20\%. We explore the effects of varying watermarking rates $\gamma \in \{10\%, 20\%, 30\%\}$ using the Pokemon dataset, while keeping all other settings unchanged. As shown in Table~\ref{watermarking_rate}, our CEAT2I remains highly effective across all tested watermarking rates, consistently outperforming other methods. Meanwhile, our CEAT2I also maintains similar performance on benign inputs.

\begin{table*}[t]
\centering
\caption{The CLIP similarity between images (CLIP \%) and watermark success rate (WSR \%) of one baseline without attacks and five different CEA methods against four types of DOV methods on the Pokemon dataset under different trigger positions. The best results among five CEA methods are highlighted in \textbf{bold}. In particular, we mark the failure cases (\textit{i.e.}, WSR $> 10\%$) among five CEA methods in \red{red}.}
\label{trigger_position}
\setlength\tabcolsep{8.6pt}{
\begin{tabular}{l|l|cc|cc|cc|cc|cc|cc}
\toprule
Trigger & \multirow{2}{*}{DOV} &  \multicolumn{2}{c|}{No Attack}  & \multicolumn{2}{c|}{ABL}  & \multicolumn{2}{c|}{NAD} & \multicolumn{2}{c|}{TPD} & \multicolumn{2}{c|}{T2IShield} & \multicolumn{2}{c}{CEAT2I (Ours)} \\ \cline{3-14}  
Position &  & CLIP & WSR  & CLIP & WSR & CLIP  & WSR & CLIP & WSR & CLIP & WSR & CLIP & WSR \\ \hline
\multirow{5}{*}{Fixed} & BadT2I-L & 89.5 & 85.6 & 78.2 & \red{81.8} & 85.7 & \red{17.8} & \textbf{89.5} & \red{90.8} & 81.2 & \red{25.8} & \textbf{89.5} & \textbf{4.6} \\
& BadT2I-G & 89.7 & 84.5 & 80.7 & \red{85.4} & 88.1 & \red{53.1} & \textbf{90.9} & \red{72.8} & 81.5 & 7.6 & 90.0 & \textbf{3.2} \\
& Rickrolling & 90.1 & 99.7 & 73.1 & \red{94.2} & 78.2 & \red{50.9} & 85.3 & \red{60.1} & 84.3 & \red{12.2} & \textbf{89.7} & \textbf{1.6} \\
& VD & 89.7 & 99.7 & 78.2 & \red{95.3} & 88.0 & \red{63.2} & 89.2 & \red{70.2} & 85.2 & \red{10.8} & \textbf{89.5} & \textbf{2.5} \\ \cline{2-14}
& Average & 89.7 & 92.4 & 77.6 & \red{89.2} & 85.0 & \red{46.3} & 88.7 & \red{73.5} & 83.1 & \red{14.1} & \textbf{89.7} & \textbf{3.0} \\ \hline
\multirow{5}{*}{Random} & BadT2I-L & 89.3 & 81.0 & 78.0 & \red{96.4} & 85.6 & \red{18.6} & \textbf{89.9} & \red{64.6} & 88.7 & \red{58.6} & 89.4 & \textbf{2.5} \\
& BadT2I-G & 89.6 & 80.1 & 80.6 & \red{89.4} & 87.9 & \red{45.0} & \textbf{90.7} & \red{73.8} & 90.6 & \red{41.7}  & 89.8 & \textbf{1.6} \\
& Rickrolling & 89.7 & 93.5 & 84.8 & \red{89.5} & 83.9 & \red{43.3} & 83.5 & \red{90.3} & 84.8 & \red{47.6}  & \textbf{88.4} & \textbf{2.9}  \\
& VD & 89.3 & 90.2 & 90.6 & \red{88.2} & 89.9 & \red{40.3} & 88.4 & \red{86.2} & \textbf{90.6} & \red{40.5}  & 88.7 & \textbf{3.1} \\ \cline{2-14}
& Average & 89.5 & 86.2 & 83.5 & \red{90.9} & 86.8 & \red{36.8} & 88.1 & \red{78.7} & 88.7 & \red{47.1} & \textbf{89.1} & \textbf{2.5} \\ 
\bottomrule
\end{tabular}}
\end{table*}

\begin{table*}[t]
\centering
\caption{The CLIP similarity between images (CLIP \%) and watermark success rate (WSR \%) of one baseline without attacks and five different CEA methods on the Pokemon dataset under a single word trigger and a phrase trigger. The best results among five CEA methods are highlighted in \textbf{bold}. In particular, we mark the failure cases (\textit{i.e.}, WSR $> 10\%$) among five CEA methods in \red{red}.}
\label{trigger_multiple_words}
\setlength\tabcolsep{6.7pt}{
\begin{tabular}{c|c|cc|cc|cc|cc|cc|cc}
\toprule
\multirow{2}{*}{Trigger}  & \multirow{2}{*}{Target Image} &  \multicolumn{2}{c|}{No Attack}  & \multicolumn{2}{c|}{ABL}  & \multicolumn{2}{c|}{NAD} & \multicolumn{2}{c|}{TPD} & \multicolumn{2}{c|}{T2IShield} & \multicolumn{2}{c}{CEAT2I (Ours)} \\ \cline{3-14}  
 &  & CLIP & WSR  & CLIP & WSR & CLIP  & WSR & CLIP & WSR & CLIP & WSR & CLIP & WSR \\ \hline
university & Local Patch & 89.5 & 85.6 & 78.2 & \red{81.8} & 85.7 & \red{17.8} & \textbf{89.5} & \red{90.8} & 81.2 & \red{25.8} & \textbf{89.5} & \textbf{4.6} \\
university & Global HelloKitty & 89.7 & 84.5 & 80.7 & \red{85.4} & 88.1 & \red{53.1} & \textbf{90.9} & \red{72.8} & 81.5 & 7.6 & 90.0 & \textbf{3.2} \\
o (U+0B66) & Global HelloKitty & 90.1 & 99.7 & 73.1 & \red{94.2} & 78.2 & \red{50.9} & 85.3 & \red{60.1} & 84.3 & \red{12.2} & \textbf{89.7} & \textbf{1.6} \\
university & Global BabyKitty & 89.7 & 99.7 & 78.2 & \red{95.3} & 88.0 & \red{63.2} & 89.2 & \red{70.2} & 85.2 & \red{10.8} & \textbf{89.5} & \textbf{2.5} \\ 
\hline
\multirow{3}{*}{\shortstack{dataset copyright \\ protection}}& Local Patch & 89.3 & 88.5 & 76.2 & \red{86.8} & 80.1 & \red{80.4} & 87.1 & \red{87.4} & 85.5 & \red{58.3} & \textbf{90.2} & \textbf{3.9} \\
& Global HelloKitty & 88.8 & 99.9 & 80.0 & \red{99.7} & 87.9 & \red{97.8} & 89.2 & \red{99.3} & 84.7 & 8.6 & \textbf{89.5} & \textbf{3.5} \\
& Global BabyKitty & 89.7 & 99.8 & \textbf{90.6} & \red{96.9} & 84.7 & \red{98.5} & 85.7 & \red{99.7} & 88.8 & \red{13.9} & 89.7 & \textbf{2.9}  \\ 
\bottomrule
\end{tabular}}
\end{table*}

\begin{table*}[t]
\centering
\caption{The computational time (h) of one baseline without attacks and five different CEA methods against BadT2I-L on Pokemon, Ossaili, and Pranked03 datasets. For the five CEA methods, the reported time cost reflects only the additional computational overhead beyond the fine-tuning stage, since model fine-tuning on watermarked datasets is required even under the no-attack setting.}
\label{computational_time_cost}
\setlength\tabcolsep{13.7pt}{
\begin{tabular}{c|ccccccccc}
\toprule
\multirow{2}{*}{Dataset} & \multirow{2}{*}{No Attack}  & \multirow{2}{*}{ABL}  & \multirow{2}{*}{NAD} & \multirow{2}{*}{TPD} & \multirow{2}{*}{T2IShield} & \multicolumn{4}{c}{CEAT2I (Ours)}  \\  
 & & & & & & Stage 1 & Stage 2 & Stage 3 & Total \\ \hline
Pokemon & 8.5 & 1.6 & 1.9 & 0.0 & 0.9 & 0.5 & 0.4 & 0.3 & 1.2 \\
Ossaili & 12.0 & 2.2 & 2.4 & 0.0 & 1.2 & 0.6 & 0.5 & 0.3 & 1.4 \\
Pranked03 & 60.0 & 11.2 & 13.5 & 0.0 & 6.1 & 3.0 & 2.8 & 1.5 & 7.3 \\
\bottomrule
\end{tabular}}
\end{table*}

\vspace{0.3em}
\noindent \textbf{Ablation on Trigger Position}.
We also investigate the impact of different trigger positions using the Pokemon dataset. By default, triggers in DOV are placed at the fixed first positions. We compare this with scenarios where trigger positions are randomized. As shown in Table~\ref{trigger_position}, our results indicate that the trigger's placement has a negligible impact on CEAT2I’s attack performance. This finding underscores that CEAT2I’s effectiveness is independent of trigger placement, maintaining the  superior performance compared to other methods in all tested scenarios of the trigger position.

\subsection{Discussions}

\noindent \textbf{Discussions on Single-Word Trigger and Phrase Trigger.} We explore the impact of both single-word and phrase triggers using the Pokemon dataset. Specifically, BadT2I-L, BadT2I-G, and VD use the single-word trigger ``university'' with different target images, while Rickrolling uses the single-character trigger ``o (U+0B66)'' in default. Additionally, we adopt the phrase ``dataset copyright protection'' as a multi-word trigger. As shown in Table~\ref{trigger_multiple_words}, CEAT2I remains highly effective for phrase triggers and achieves the lowest WSRs compared to baselines. This is primarily because trigger words dominate the learned features, and even when the trigger appears as a phrase, each constituent word behaves as a statistical outlier. Consequently, the proposed outlier-based detection in the second stage can effectively capture such phrase triggers.

\vspace{0.3em}
\noindent \textbf{Discussions on Computational Cost.} In Table~\ref{computational_time_cost}, we report the computational time (in hours) of one baseline without attacks and five different CEA methods against BadT2I-L across three datasets. All the experiments are conducted on one NVIDIA 4090 GPU. For the five CEA methods, the reported time cost represents only the additional computational overhead beyond the mandatory fine-tuning stage, since fine-tuning on watermarked datasets is required even in the no-attack setting. As shown in this table, among these methods, TPD incurs almost no additional time cost, as it only applies random perturbations to input texts during inference. In addition, CEAT2I requires similar time as T2IShield but less than ABL and NAD, as both ABL and NAD involve additional fine-tuning of the watermarked models. Notably, CEAT2I achieves the lowest WSRs among all compared approaches, demonstrating its superior effectiveness. In the future work, we plan to investigate strategies for further reducing the computational overhead of CEAs while preserving its effectiveness.



\vspace{0.3em}
\noindent \textbf{Discussions on Watermarks for Style Protection.} In addition to the backdoor-based DOV, we evaluate the robustness of a representative watermarking approach for style protection, SIREN~\cite{li2024towards}, under our CEAT2I.
SIREN embeds an imperceptible but learnable coating into protected fine-tuning datasets so that personalized diffusion models can reliably capture it during the fine-tuning process. For verification under black-box conditions, features of generated outputs are extracted and classified to determine the presence of the coated signature.

Following SIREN, we conduct experiments on five datasets, including Pokemon~\cite{pokemon}, CelebA-HQ~\cite{karras2018progressive}, ArtBench~\cite{liao2022artbench}, Landscape~\cite{kaggle2024landscape}, and WikiArt~\cite{saleh2015large} datasets.
Unless otherwise specified, the fine-tuning and evaluation configurations follow the original SIREN paper. For robustness evaluation, we use a baseline without attacks and our CEAT2I described in our original manuscript under the same settings. Effectiveness is evaluated using Bit Accuracy (BitAcc) for verification and CLIP scores for generation quality. The higher BitAcc and CLIP scores indicate better reliability of the method.

We compare the BitAcc and CLIP scores of a baseline without attacks and our CEAT2I against SIREN across five datasets, as reported in Table~\ref{fig:siren}. In the absence of attacks, SIREN attains high BitAcc for verification while preserving strong generation quality on benign images. However, SIREN suffers significant performance degradation under our CEAT2I, as evidenced by notably reduced BitAcc and CLIP scores.
These results highlight that our proposed CEAT2I effectively targets both harmful backdoor-based watermarking methods (BadT2I-G, Rickrolling, and VD) and harmless DOV schemes (BadT2I-L and SIREN). 

\subsection{Resistance to Potential Adaptive Defense}
In the previous experiments, we assume that the data owner is unaware of the CEAT2I attack. In this section, we consider a more challenging setting, where the data owner knows the existence of CEAT2I and generates the watermarked samples with an adaptive defense. Recall that CEAT2I detects watermarked samples by measuring the feature deviation between the original and fine-tuned T2I diffusion models. Therefore, an effective adaptive defense would aim to minimize this feature deviation during watermark insertion, making watermarked samples harder to detect. To achieve this adaptive defense, the data owner first trains a T2I diffusion model on the benign datasets. Then, they optimize a universal textual trigger specifically to reduce the feature deviation during fine-tuning. This is done using a discrete optimization process~\cite{yang2024multi} over the token space. Concretely, we search for a 4-token trigger appended to benign prompts, which introduces the minimal difference between the original and fine-tuned model representations.

\begin{table}[t]
\centering
\caption{The CLIP similarity between images (CLIP \%) and Bit Accuracy (BitAcc \%) of one baseline without attacks and our CEAT2I against SIREN on five datasets, including Pokemon, CelebA-HQ, ArtBench, Landscape, WikiArt.}
\label{fig:siren}
\setlength\tabcolsep{13pt}{
\begin{tabular}{l|cc|cc}
\toprule
\multirow{2}{*}{Dataset} & \multicolumn{2}{c|}{No Attack} & \multicolumn{2}{c}{CEAT2I (Ours)} \\
& CLIP & BitAcc & CLIP & BitAcc \\ \hline
Pokemon & 88.2 & 94.5 & 86.4 & 0.0 \\
CelebA-HQ & 86.8 & 95.0 & 88.5 & 0.0 \\
ArtBench & 87.5 & 92.4 & 85.0 & 0.0 \\
Landscape & 89.0 & 96.5 & 86.4 & 0.0 \\ 
WikiArt & 85.5 & 90.0 & 83.6 & 0.0 \\ 
\bottomrule
\end{tabular}}
\end{table}

\orange{Let $f_{\boldsymbol{\theta}}^i\left(\boldsymbol{z}_t, t, \boldsymbol{c}\right)$ and $f_{\boldsymbol{\theta}_{\boldsymbol{w}_t}'}^i\left(\boldsymbol{z}_t, t, \boldsymbol{c}\right)$ denote the feature activations at the $i$-th layer of the original and fine-tuned diffusion models at an early fine-tuning epoch $T_{e}$, respectively. For an image $\boldsymbol{x}$ and text input $y$, we define $\boldsymbol{z}_t=\mathcal{E}(\boldsymbol{x})$ as the latent representation at diffusion timestep $t$, and $\boldsymbol{c}=\mathcal{T}(y)$ as the semantic text embedding. Let $N$ be the total number of layers in the diffusion model. We denote benign image–text pairs by $(\boldsymbol{x}_b, y_b) \in \mathcal{D}_{b}$, and watermarked pairs by $(\boldsymbol{x}_w, y_w) \in \mathcal{D}_{w}$, where $\boldsymbol{x}_w$ is the target image and the watermarked prompt is $y_{w} = y \oplus p$, with $p$ representing the trigger tokens. In the adaptive attack, the attacker optimizes a trigger $p$ such that the watermarked prompt $y_w$ produces a target image $\boldsymbol{x}_w$ while minimizing the feature deviation of the watermarked and benign samples. This objective can be formalized as follows:}
\begin{equation}
\begin{aligned}
\orange{\min_{p} \frac{1}{N|\mathcal{D}_w|} \sum_{(\boldsymbol{x}_w, y_w) \in \mathcal{D}_w}  \sum_{i=1}^N
||f_{\boldsymbol{\theta}}^i\left(\mathcal{E}(\boldsymbol{x}_w), t, \mathcal{T}(y_w)\right)}\\\orange{-f_{\boldsymbol{\theta}_{\boldsymbol{w}_t}'}^i\left(\mathcal{E}(\boldsymbol{x}_w), t, \mathcal{T}(y_w)\right)||_2^2} \\ \orange{- \frac{1}{N|\mathcal{D}_b|} \sum_{(\boldsymbol{x}_b, y_b) \in \mathcal{D}_b}  \sum_{i=1}^N
||f_{\boldsymbol{\theta}}^i\left(\mathcal{E}(\boldsymbol{x}_b), t, \mathcal{T}(y_b)\right)}\\\orange{-f_{\boldsymbol{\theta}_{\boldsymbol{w}_t}'}^i\left(\mathcal{E}(\boldsymbol{x}_b), t, \mathcal{T}(y_b)\right)||_2^2.}
\label{eq:adaptive_attack}
\end{aligned}
\end{equation}

We conduct this experiment on the Pokemon dataset, using 10,000 optimization steps with a learning rate of 0.001. The optimized trigger achieves a CLIP score of 88.8\% and a WSR of 97.8\% when no attack is applied. It indicates that the watermark is both stealthy and effective under standard conditions. However, when applying our CEAT2I against this adaptive defense, we observe a CLIP score of 89.2\% and a WSR of only 3.4\%, meaning that our method can still successfully remove the watermark without harming benign generation quality. This demonstrates that our CEAT2I remains effective even in the face of adaptive defenses. 
The probable reason is that the trigger pattern is optimized on the surrogate model and has low transferability, highlighting the robustness and practicality of CEAT2I in more adversarial settings.

\section{Potential Limitations and Future Directions}

As the first work to explore CEA against DOV for T2I diffusion models, our CEAT2I inevitably has some limitations.

Firstly, although CEAT2I does not require any additional fine-tuning beyond the standard model fine-tuning on watermarked datasets, it introduces extra computational overhead during the watermark removal process. Specifically, it relies on extracting intermediate feature representations to detect watermarked samples and identify triggers, which adds additional time and resource consumption. A promising direction for future research is to further simplify the CEAT2I pipeline. The goal of future work could be to develop an end-to-end framework that automatically integrates detection, identification, and mitigation into a single lightweight process. 

Secondly, our CEAT2I is designed specifically for T2I models, such as Stable Diffusion, which rely on the alignment between textual prompts and visual content. While these models currently dominate the generative image synthesis landscape, the broader generative AI ecosystem is rapidly evolving to include other modalities, such as text-to-video, text-to-3D, and text-image-language foundation models. In these settings, the architecture and modality  differ significantly. The effectiveness of CEAT2I has not been validated outside the image generation tasks. As such, a key direction for future work is to explore whether the foundational ideas behind CEAT2I, such as early convergence analysis and concept erasure, can be extended for other multimodal generative models.

Finally, it is important to note that while our proposed CEAT2I demonstrates the feasibility of undermining current backdoor-based DOV schemes, its existence calls for stronger, more secure DOV methods. Future work should not only focus on improving attack techniques but also inspire the community to design more robust DOV  methods that are resistant to CEA like our proposed CEAT2I.

\section{Conclusion}
In this paper, we presented CEAT2I, a novel and effective copyright evasion attack targeting DOV in T2I models. While DOV techniques offered a promising solution for protecting datasets via backdoor-based watermarking, we demonstrated that they remain vulnerable to well-crafted evasion attacks. Our method leveraged three key components, including watermarked sample detection via feature convergence analysis, trigger identification through token-level ablation, and efficient watermark removal via closed-form model editing. Extensive experiments across four DOV methods and three datasets showed that our CEAT2I significantly outperformed prior potential attack methods, effectively removing watermarks while preserving model fidelity and visual quality. 

\noindent \textbf{Ethics Statement.} This work aims to investigate the security vulnerabilities of DOV methods based on backdoor techniques in T2I diffusion models. All experiments with our proposed CEAT2I are conducted strictly within controlled laboratory environments, using only publicly available open-source datasets. We  emphasize that CEAT2I is designed solely for research purposes to highlight potential risks in existing DOV mechanisms. We do not support the deployment of CEAT2I in real-world applications for malicious purposes. 

\section*{Acknowledgement}
This work is supported in part by the National Natural Science Foundation of China under Grant 62571298 and Tsinghua SIGS KA Cooperation Fund.

\bibliographystyle{plain}
\bibliography{egbib}

\appendix

In our CEAT2I, the watermarked sample detection uses all layers of the U-net to calculate feature deviations following~\cite{basu2024localizing}. To account for inter-layer scale differences, we normalize the deviation scores per layer in U-net as suggested in~\cite{basu2024localizing}. Then, we apply a voting mechanism across layers, which reduces the impact of anomalies in any single layer. Since the lower layers capture the low-level pixel information and the deeper layers capture the semantic information, this full-layer aggregation for watermarked sample detection fully adopts both feature information of all layers, which can ensure detection consistency and robustness. The adopted layers are listed in Table~\ref{fig:down-block-layer}, Table~\ref{fig:mid-block-layer}, and Table~\ref{fig:up-block-layer}.

\begin{table}[htbp]
  \centering
 \caption{Layer Mappings for the Down-Block in the UNet.}
  \resizebox{\columnwidth}{!}{\begin{tabular}{llll}
    \toprule
    Layer    & Type of Layer  & Layer Name \\  
    \midrule
    0   & self-attention & down-blocks.0.attentions.0.transformer-blocks.0.attn1 \\
    1  & cross-attention & down-blocks.0.attentions.0.transformer-blocks.0.attn2 \\
    2    & feedforward &  down-blocks.0.attentions.0.transformer-blocks.0.ff \\
    3   & self-attention & down-blocks.0.attentions.1.transformer-blocks.0.attn1 \\
    4     & cross-attention  & down-blocks.0.attentions.1.transformer-blocks.0.attn2 \\
    5  & feedforward & down-blocks.0.attentions.1.transformer-blocks.0.ff  \\
    6  & self-attention & down-blocks.0.resnets.0  \\
    7  & resnet & down-blocks.0.resnets.1  \\
    8  & self-attention & down-blocks.1.attentions.0.transformer-blocks.0.attn1  \\
    9  & cross-attention & down-blocks.1.attentions.0.transformer-blocks.0.attn2  \\
    10  & feedforward & down-blocks.1.attentions.0.transformer-blocks.0.ff  \\
    11  & self-attention & down-blocks.1.attentions.1.transformer-blocks.0.attn1  \\
    12  & cross-attention & down-blocks.1.attentions.1.transformer-blocks.0.attn2 \\
    13  & feedforward & down-blocks.1.attentions.1.transformer-blocks.0.ff  \\
    14  & resnet & down-blocks.1.resnets.0  \\
    15  & resnet & down-blocks.1.resnets.1  \\
    16  & self-attention & down-blocks.2.attentions.0.transformer-blocks.0.attn1  \\
    17  & cross-attention & down-blocks.2.attentions.0.transformer-blocks.0.attn2  \\
    18  & feedforward & down-blocks.2.attentions.0.transformer-blocks.0.ff  \\
    19  & self-attention & down-blocks.2.attentions.1.transformer-blocks.0.attn1 \\
    20  & cross-attention & down-blocks.2.attentions.1.transformer-blocks.0.attn2  \\
    21  & feedforward & down-blocks.2.attentions.1.transformer-blocks.0.ff  \\
    22  & resnet & down-blocks.2.resnets.0  \\
    23  & resnet & down-blocks.2.resnets.1  \\
    24  & resnet & down-blocks.3.resnets.0  \\
    25  & resnet & down-blocks.3.resnets.1  \\
    \bottomrule
  \end{tabular}}
 \label{fig:down-block-layer}
\end{table}

\begin{table}[htbp]
  \centering
 \caption{Layer Mappings for the Mid-Block in the UNet.}
  \resizebox{\columnwidth}{!}{\begin{tabular}{llll}
    \toprule
    Layer    & Type of Layer  & Layer Name \\  
    \midrule
    0   & self-attention & mid-block.attentions.0.transformer-blocks.0.attn1 \\
    1  & cross-attention & mid-block.attentions.0.transformer-blocks.0.attn2 \\
    2    & feedforward &  mid-block.attentions.0.transformer-blocks.0.ff \\
    3   & resnet & mid-block.resnets.0 \\
    4     & resnet  & mid-block.resnets.1 \\
    \bottomrule
  \end{tabular}}
 \label{fig:mid-block-layer}
\end{table}

\begin{table}[htbp]
  \centering
 \caption{Layer Mappings for the Up-Block in the UNet.}
  \resizebox{\columnwidth}{!}{\begin{tabular}{llll}
    \toprule
    Layer    & Type of Layer  & Layer Name \\  
    \midrule
    0   & resnet & up-blocks.0.resnets.0 \\
    1  & resnet & up-blocks.0.resnets.1 \\
    2    & resnet &  up-blocks.0.resnets.2 \\
    3   & self-attention & up-blocks.1.attentions.0.transformer-blocks.0.attn1 \\
    4     & cross-attention  & up-blocks.1.attentions.0.transformer-blocks.0.attn2 \\
    5  & feedforward & up-blocks.1.attentions.0.transformer-blocks.0.ff  \\
    6  & self-attention & up-blocks.1.attentions.1.transformer-blocks.0.attn1  \\
    7  & cross-attention & up-blocks.1.attentions.1.transformer-blocks.0.attn2  \\
    8  & feedforward & up-blocks.1.attentions.1.transformer-blocks.0.ff  \\
    9  & self-attention & up-blocks.1.attentions.2.transformer-blocks.0.attn1  \\
    10  & cross-attention & up-blocks.1.attentions.2.transformer-blocks.0.attn2  \\
    11  & feedforward & up-blocks.1.attentions.2.transformer-blocks.0.ff  \\
    12  & resnet & up-blocks.1.resnets.0 \\
    13  & resnet & up-blocks.1.resnets.1  \\
    14  & resnet & up-blocks.1.resnets.2  \\
    15  & self-attention & up-blocks.2.attentions.0.transformer-blocks.0.attn1  \\
    16  & cross-attention & up-blocks.2.attentions.0.transformer-blocks.0.attn2  \\
    17  & feedforward & up-blocks.2.attentions.0.transformer-blocks.0.ff  \\
    18  & self-attention & up-blocks.2.attentions.1.transformer-blocks.0.attn1  \\
    19  & cross-attention & up-blocks.2.attentions.1.transformer-blocks.0.attn2 \\
    20  & feedforward & up-blocks.2.attentions.1.transformer-blocks.0.ff  \\
    21  & self-attention & up-blocks.2.attentions.2.transformer-blocks.0.attn1  \\
    22  & cross-attention & up-blocks.2.attentions.2.transformer-blocks.0.attn2  \\
    23  & feedforward & up-blocks.2.attentions.2.transformer-blocks.0.ff  \\
    24  & resnet & up-blocks.2.resnets.0  \\
    25  & resnet & up-blocks.2.resnets.1  \\
    26  & resnet & up-blocks.2.resnets.2  \\
    27  & self-attention & up-blocks.3.attentions.0.transformer-blocks.0.attn1  \\
    28  & cross-attention & up-blocks.3.attentions.0.transformer-blocks.0.attn2  \\
    29  & feedforward & up-blocks.3.attentions.0.transformer-blocks.0.ff  \\
    30  & self-attention & up-blocks.3.attentions.1.transformer-blocks.0.attn1  \\
    31  & cross-attention & up-blocks.3.attentions.1.transformer-blocks.0.attn2  \\
    32  & feedforward & up-blocks.3.attentions.1.transformer-blocks.0.ff  \\
    33  & self-attention & up-blocks.3.attentions.2.transformer-blocks.0.attn1  \\
    34  & cross-attention & up-blocks.3.attentions.2.transformer-blocks.0.attn2 \\
    35  & feedforward & up-blocks.3.attentions.2.transformer-blocks.0.ff  \\
    36  & resnet & up-blocks.3.resnets.0  \\
    37  & resnet & up-blocks.3.resnets.1  \\
    38  & resnet & up-blocks.3.resnets.2  \\
    \bottomrule
  \end{tabular}}
 \label{fig:up-block-layer}
\end{table}

\end{document}